\documentclass[final]{cvpr}

\usepackage{times}
\usepackage{epsfig}
\usepackage{graphicx}
\usepackage{amsmath}
\usepackage{amssymb}
\usepackage{multirow}
\usepackage{caption}
\usepackage{bbm}
\usepackage{cases}
\usepackage{booktabs}
\usepackage{dsfont}
\usepackage{diagbox}
\usepackage{float}
\usepackage{tabularx} %

\usepackage{amsmath,amsfonts,bm}

\def\eqref#1{equation~\ref{#1}}

\def\1{\bm{1}}

\def\rvx{{\mathbf{x}}}

\def\rvz{{\mathbf{z}}}

\DeclareMathAlphabet{\mathsfit}{\encodingdefault}{\sfdefault}{m}{sl}
\SetMathAlphabet{\mathsfit}{bold}{\encodingdefault}{\sfdefault}{bx}{n}

\newcommand{\R}{\mathbb{R}}

\newcommand{\softmax}{\sigma_{\mathrm{SM}}}
\newcommand{\length}{\mathrm{length}}

\DeclareMathOperator{\conf}{conf}
\DeclareMathOperator{\acc}{acc}

\usepackage[super]{nth}
\usepackage[square,numbers]{natbib}

\usepackage[pagebackref=true,breaklinks=true,colorlinks,bookmarks=false]{hyperref}

\def\cvprPaperID{4650} %
\def\confYear{CVPR 2021}

\begin{document}

\title{On Calibration of Scene-Text Recognition Models}

\author{

    Ron Slossberg$^1$, Oron Anschel$^2$, Amir Markovitz$^2$, Ron Litman$^2$, Aviad Aberdam$^1$, Shahar Tsiper$^2$, \\
    Shai Mazor$^2$, Jon Wu$^2$ and R. Manmatha$^2$\\
    $^1$Technion, $^2$Amazon Web Services\\
    {\tt\small \{ronslos,  aaberdam\}@campus.technion.ac.il}\\
    {\tt\small \{oronans, amirmak, litmanr, tsiper, smazor, jonwu, manmatha\}@amazon.com}
    }

\maketitle

\definecolor{antiquefuchsia}{rgb}{0.57, 0.36, 0.51}  %

\ifdefined\ShowNotes
  \newcommand{\colornote}[3]{{\color{#1}\bf{#2: #3}\normalfont}}
\else
  \newcommand{\colornote}[3]{}
\fi

\newcommand {\rons}[1]{\colornote{blue}{RonS}{#1}}
\newcommand {\ronl}[1]{\colornote{green}{RonL}{#1}}
\newcommand {\oron}[1]{\colornote{red}{Oron}{#1}}
\newcommand {\shahar}[1]{\colornote{magenta}{Shahar}{#1}}
\newcommand {\amir}[1]{\colornote{antiquefuchsia}{Amir}{#1}}

\newcommand {\shai}[1]{\colornote{red}{Shai}{#1}}
\newcommand {\manmatha}[1]{\colornote{red}{Manmatha}{#1}}

\begin{abstract}

In this work, we study the problem of word-level confidence calibration for scene-text recognition (STR).
Although the topic of confidence calibration has been an active research area for the last several decades, the case of structured and sequence prediction calibration has been scarcely explored.
We analyze several recent STR methods and show that they are consistently overconfident. We then focus on the calibration of STR models on the word rather than the character level. 
In particular, we demonstrate that for attention based decoders, calibration of individual character predictions increases word-level calibration error compared to an uncalibrated model.
In addition, we apply existing calibration methodologies as well as new sequence-based extensions to numerous STR models, demonstrating reduced calibration error by up to a factor of nearly $7$.
Finally, we show consistently improved accuracy results by applying our proposed sequence calibration method as a preprocessing step to beam-search.

\end{abstract}

\section{Introduction}

Scene Text Recognition (STR) -- the task of extracting text from a cropped word image, has seen an increase in popularity in recent years.
While an active research area for almost three decades, STR performance has just recently seen a significant performance boost due to the utilization of deep-learning models~\cite{shi2016end, Baek2019clova, Bai2018aster, bartz2019kiss, qiao2020seed, Litman_2020_CVPR}.
Some typical applications relying on STR models include assistance to the visually impaired, content moderation in social media, automated processing of passports, and street sign recognition for autonomous vehicles.

The above examples, often referred to as user-facing applications, require precise estimates of the prediction confidence \ie what the probability for a correct prediction is.
For instance, an incorrect prediction made by an STR model embedded in an autonomous driving system could endanger humans in the vicinity. At the same time, a correct confidence assessment might mitigate the damage.

\begin{figure}[t]
\begin{center}
  \includegraphics[width=0.75\linewidth]{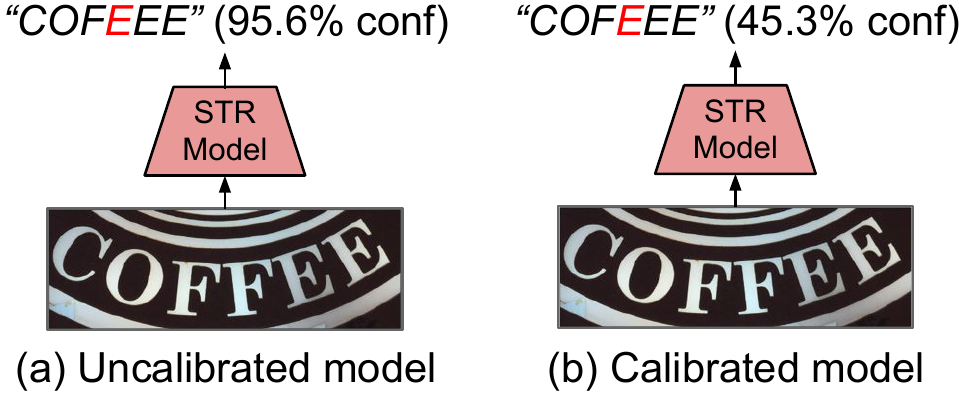}
\end{center}
    \vspace{-0.2cm}
   \caption{\textbf{Uncalibrated vs. Calibrated Model}. Uncalibrated models often overestimate their confidence.}
\label{fig:calib_samples}
  \vspace{-0.2cm}
\end{figure}
\begin{figure}[t]
\begin{center}
  \includegraphics[width=0.75\linewidth]{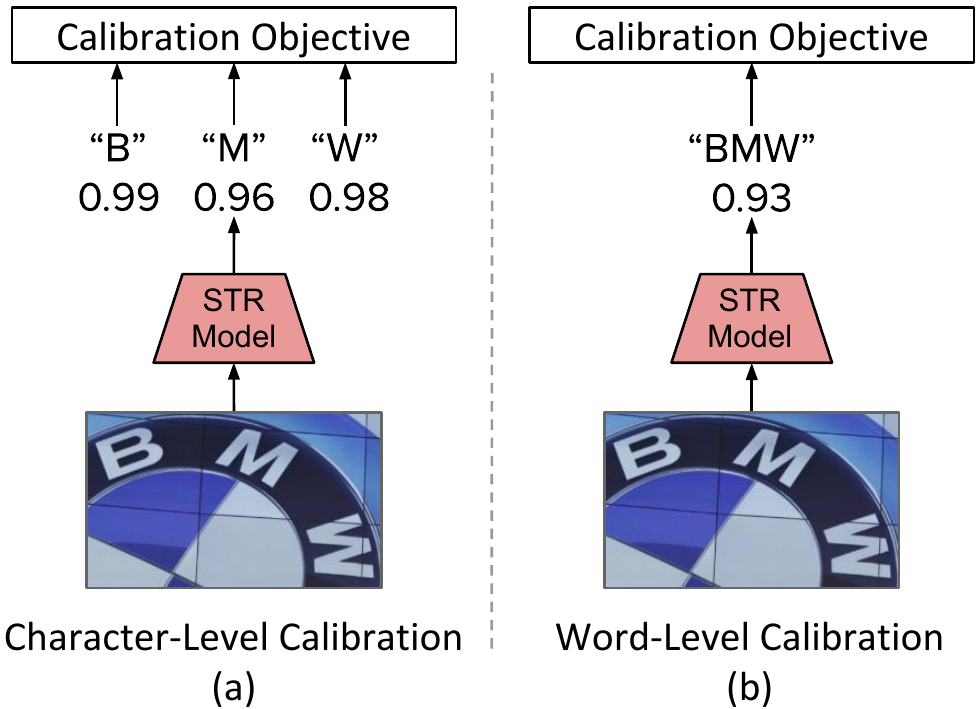}
\end{center}
\vspace{-0.1cm}
   \caption{\textbf{Character vs. Word-Level Calibration}. (a) Individual character score calibration. (b) Word-level calibration \ie the confidence score for the entire word is directly optimized. We demonstrate the importance of adopting (b) as opposed to (a).}
\label{fig:optimzation}
\vspace{-0.1cm}
\end{figure}

Confidence calibration is the task of tuning the model confidence scores to match a successful prediction's underlying probability.
For example, within the group of samples producing a confidence score of $0.7$, we expect to achieve a prediction success rate of exactly $70\%$.
In mathematical terms perfect calibration is defined as:
\begin{equation}
\mathbb{P}(\hat{Y}=Y \mid \hat{P}=p)=p, \quad \forall p \in[0,1],
\label{eq:calibration}
\end{equation}
where $\hat{Y},Y$ are the prediction and ground-truth labels respectively and $\hat{P}$ is the estimated confidence score.

Confidence calibration and model reliability have been active areas of research for many years~\cite{brier1950verification, degroot1983comparison, niculescu2005predicting}; however, the task of calibrating sequence-level confidence has received little attention and
to the best of our knowledge, in STR, it has yet to be explored.
Previous work studied calibration, and confidence intervals for structured prediction in the context of NLP problems ~\cite{kumar2019calibration, desai2020calibration,Nguyen2015a,leathart2020temporal}. 
These methods focus on the calibration of marginal confidences, analogous to the calibration of each decoding step or each token, as depicted in Figure~\ref{fig:optimzation}~(a).
This work, on the other hand, advocates for direct calibration of the word-level scalar confidence scores, as illustrated in Figure ~\ref{fig:optimzation}~(b), and motivates this by theoretical reasoning and an empirical evaluation.

Kuleshov and Liang~\cite{Kuleshov2015} laid the mathematical framework for confidence calibration in structured prediction problems.
In their framework, they define a sequence-level objective for the calibration score. 
In their paper, they state that \textit{"If a user only looked at the marginals for the first position, she might be sorely disappointed."}.
For example, in our context, this could mean calibrating each decoding step.
Here we utilize the framework proposed by Kuleshov and Liang~\cite{Kuleshov2015}, extending their methodologies to modern scene-text recognition architectures.

\begin{figure*}[t!]
\begin{center}
    \vspace{-4pt}
  \includegraphics[width=0.26\textwidth]{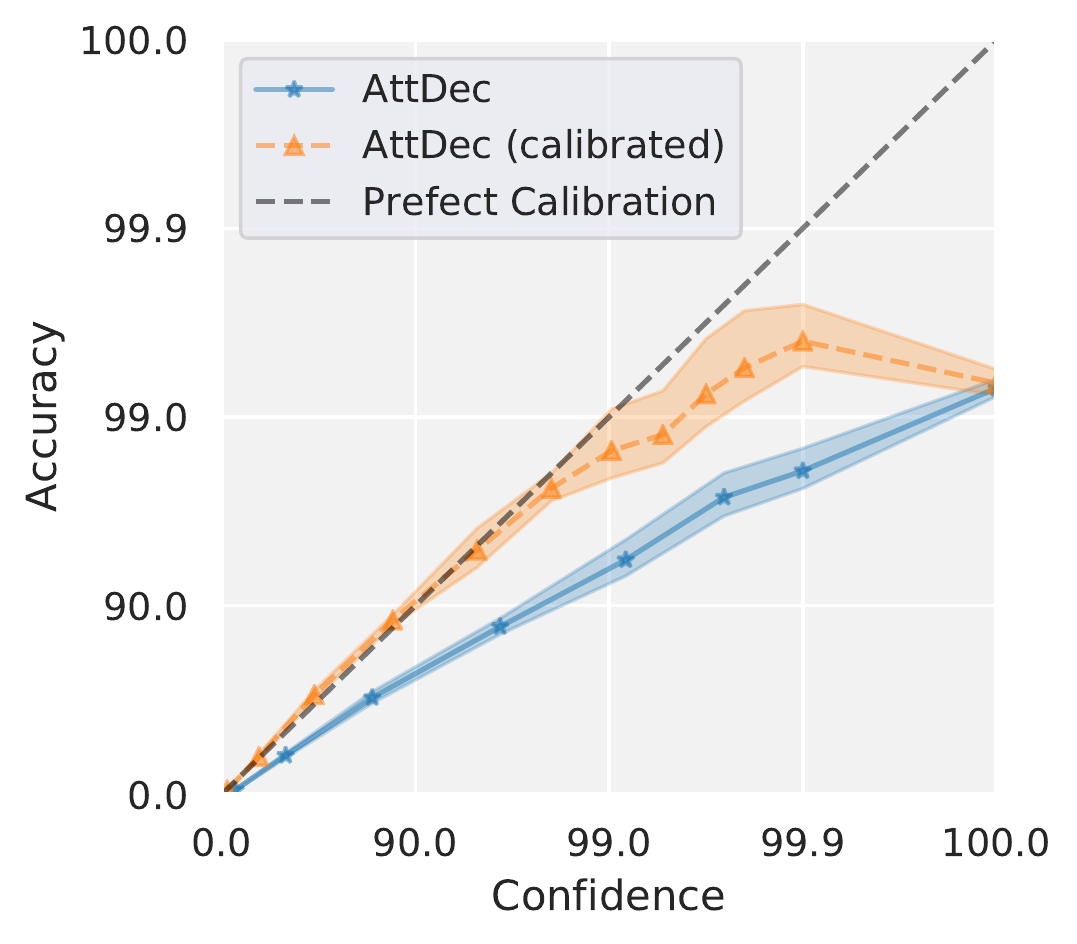}
  \hspace{-12pt}
  \includegraphics[width=0.26\textwidth]{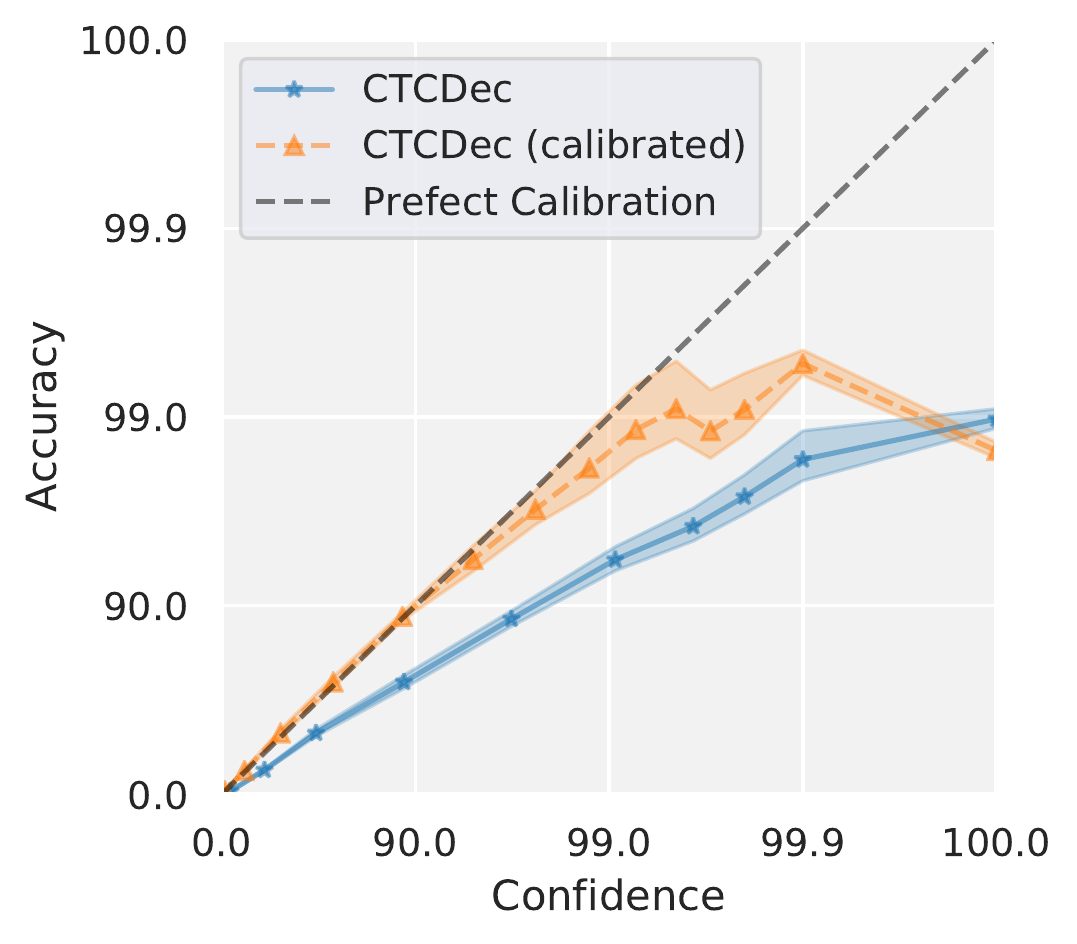}
    \hspace{-12pt}
  \includegraphics[width=0.26\textwidth]{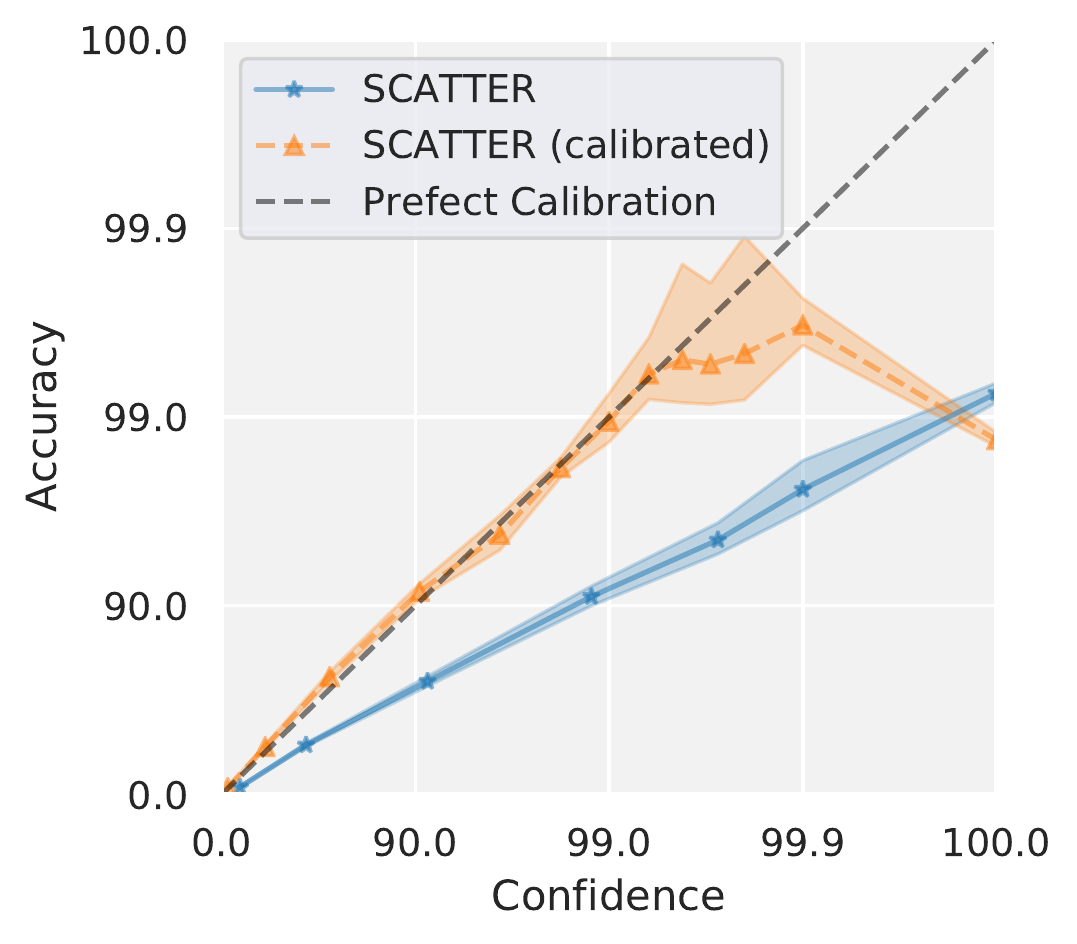}
    \hspace{-12pt}
  \includegraphics[width=0.26 \textwidth]{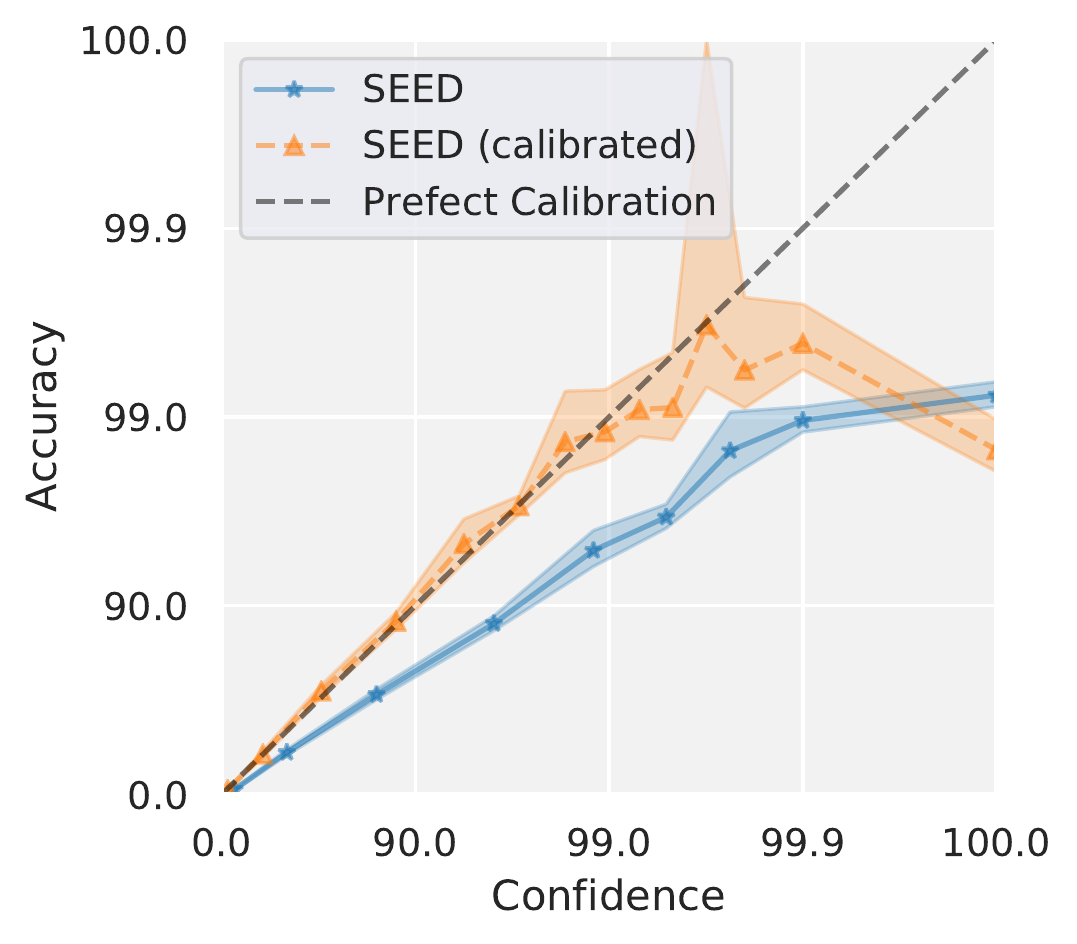}
    \hspace{-12pt}
    \vspace{-10pt}
\end{center}
   \caption{\textbf{Reliability Diagrams \cite{degroot1983comparison}}: 
   (i) AttDec --  a variant of ~\cite{Baek2019clova} with an attention decoder, 
   (ii) CTCDec -- a variant of ~\cite{Baek2019clova} with a CTC decoder, 
   (iii) SCATTER \cite{Litman_2020_CVPR} and 
   (iv) SEED \cite{qiao2020seed}. We calibrate using T-scaling coupled with an equal bin size ECE objective applied to the word-level scalar confidence scores. The accuracy here is measured w.r.t exact word match.
   The figure shows accuracy vs. confidence plotted for equally-sized confidence bins, before and after calibration. Over-confidence can be observed for STR models, where the confidence of the model is higher than the expected accuracy.
  }
\vspace{-0.3cm}
\label{fig:reliability_diagrams}
\end{figure*}

This work studies the confidence characteristics of STR models and confirms the overconfidence tendency displayed by other DNN models, where the confidence score is higher than the empirical accuracy
~\cite{on_calib_icml17,lakshminarayanan2017simple, hendrycks2016baseline, pereyra2017regularizing} (see Figures~\ref{fig:calib_samples}~and~\ref{fig:reliability_diagrams}). To conduct a comprehensive study, we re-implemented several recent STR methods and trained them under the same conditions (see supplementary for accuracy results).

We proceed to propose new methodologies for word-level calibration of pre-trained STR models, adapting existing calibration methods, namely Temperature-scaling (T-scaling) \cite{on_calib_icml17, platt1999probabilistic}, to this task.
Surprisingly, we show that character-level calibration of STR models based on transformer~\cite{Vaswani2017trans} or attention~\cite{li2019show,Bai2018aster} decoders, \emph{adversely affects} the word-level calibration error relative to a non-calibrated baseline.

In addition, we propose a sequence oriented extension to Temperature-scaling named Step Dependent T-Scaling. Furthermore, we extend the Expected Calibration Error (ECE) \cite{naeini2015obtaining} proposed for binary classification problems to the regime of sequence calibration by incorporating a sequence accuracy measure, namely the edit-distance~\cite{levenshtein1966binary} metric. 
Finally, we present a useful application for confidence calibration by combining calibration with a beam-search decoding scheme, achieving consistent accuracy gains.
Our key contributions are as follows:
\begin{itemize}%
    \item We highlight the importance of directly calibrating for the word-level confidence score by demonstrating that performing character-level optimization often has an adverse effect on word-level calibration error.
    \item We present the first analysis of confidence estimation and calibration for STR methods.
    \item We propose T-scaling and ECE extensions suited for sequence-level calibration.
    \item We demonstrate consistent accuracy gains by applying beam-search to calibrated STR models.
\end{itemize}

\section{Related Work}
\label{sec:background}
\paragraph{Scene Text Recognition}
Shi~\etal~\cite{shi2016end}  proposed an end-to-end image to sequence approach without the need for character-level annotations. The authors used a BiLSTM~\cite{graves2013speech} for modeling contextual dependencies, and Connectionist Temporal Classification (CTC)~\cite{Graves2006ctc} for decoding.
Baek~\etal~\cite{Baek2019clova} proposed a four-stage framework unifying several previous techniques~\cite{fedor2018rosetta, shi2016end, bai2016tps, Liu2016STARNetAS}. The framework comprises the following building blocks: image transformation, feature extraction, sequence modeling, and decoding. Numerous subsequent methods also conform to this general structure~\cite{Litman_2020_CVPR,Bai2018aster,qiao2020seed,bartz2019kiss}. Currently, SOTA results are often achieved by methods adopting an attention-based~\cite{bahdanau2014neural} decoder scheme. The attention-based decoders usually consist of an RNN cell taking at each step the previously predicted token and a hidden state as inputs and outputting the next token prediction.

\paragraph{Confidence Calibration}
Model calibration has been a subject of interest within the data modeling and general scientific communities for many decades~\cite{brier1950verification, degroot1983comparison, niculescu2005predicting}.
Several recent papers \cite{on_calib_icml17,lakshminarayanan2017simple, hendrycks2016baseline, pereyra2017regularizing, ji2019bin, Nixon_2019_CVPR_Workshops} have studied model calibration in the context of modern neural networks and classifier calibration (scalar or multi-class predictions). Empirically, modern neural networks are poorly calibrated and tend towards overconfidence.
A common theme among numerous calibration papers is that Temperature-scaling (T-scaling)~\cite{on_calib_icml17} is often the most effective calibration method even when compared to complex methods such as Monte-Carlo Dropout, Deep-ensemble, and Bayesian methods \cite{ovadia2019can}.
Nixon~\etal~\cite{Nixon_2019_CVPR_Workshops} conduct a study on several proposed variations on established calibration metrics and suggest some good practices for calibration optimization and evaluation. Similarly, we minimize an ECE calibration objective using a gradient framework.

\paragraph{Confidence Calibration for Sequential Models}

Most of the confidence calibration literature is focused on calibrating a single output classifier. Kuleshov and Liang~\cite{Kuleshov2015} were the first to propose a calibration framework for structured prediction problems. The framework defines the notion of ``Events of Interest'' coupled with confidence scores allowing event-level calibration. The practical methods laid out by Kuleshov and Liang~\cite{Kuleshov2015}, however, predate the recent advances in DNNs.

Kumar~\etal~\cite{kumar2019calibration} address the problem of miscalibration in neural machine translation (NMT) systems.
The authors show that NMT models are poorly calibrated and propose a calibration method based on a T-scaling variant where the temperature is predicted at each decoding step. They were also able to improve translation performance by applying beam-search to calibrated models. Our experiments find this to be beneficial for the task of STR as well.

Desai~\etal~\cite{desai2020calibration} suggest the usage of T-Scaling for calibration of pre-trained transformers models (\eg BERT \cite{devlin2018bert}). the authors differentiate between in and out of domain calibration and propose using T-Scaling, and label-smoothing \cite{pereyra2017regularizing} techniques. We point-out that label smoothing is carried out during the training phase and therefore affects the model accuracy. Here, calibration is also conducted at the individual output level.

In another work~\cite{leathart2020temporal}, a proposed extension to T-scaling calibration for sequences is presented. The authors employ a parametric decaying exponential to model the temperature for each decoding step.
Again, similarly to \cite{kumar2019calibration} calibration is performed for each decoding step and not for entire sequences.

\section{Background}
\label{sec:background}
\paragraph{Temperature Scaling}
T-Scaling~\cite{on_calib_icml17}, a limited version of Platt Scaling ~\cite{platt1999probabilistic}, is a simple yet effective calibration method. T-scaling utilizes a single parameter $T>0$. Given a logits vector $\rvz_i$ the model produces a calibrated score as: 
\begin{equation}
    \hat q_i = \max_k ~ \softmax \left( \rvz_{i} / T \right)^{(k)},
    \label{eq:temp_scaling}
\end{equation}
where $\hat q_i$ denotes the estimated confidence of the $\text{i}^{th}$ data sample, $\rvz_{i} \in \R^K$ is the output logits and $K$ is the number of output classes (number of supported symbols for a STR model). $T$ is a global scaling parameter and $\sigma_{SM}$ is the softmax function defined as:
\begin{equation}
\softmax\left(\mathbf{\rvz}_{i}\right)^{(k)}=\frac{\exp \left(\rvz_{i}^{(k)}\right)}{\sum_{l=1}^{K} \exp \left(\rvz_{i}^{(l)}\right)}. 
\end{equation}

The temperature parameter $T$ scales the logits, either altering the predicted confidence scores as necessary.
T-Scaling is a monotonic transform of the confidence values, and therefore \emph{does not affect the classifier model accuracy}.
 
\paragraph{Reliability Diagrams} Figure~\ref{fig:reliability_diagrams} presents a visual representation of model calibration ~\cite{degroot1983comparison, niculescu2005predicting}. Reliability diagrams show the expected accuracy for different confidence bins, where the diagonal represents a perfect calibration. Within each plot, the lower right triangle represents the overconfidence regime, where the estimated sample confidence is higher than its expected accuracy. We observe that the uncalibrated models are overconfident. We note that these plots do not contain the number of bin samples, and therefore, calibration error and accuracy cannot be directly derived from them.

\paragraph{Expected Calibration Error (ECE)}

Expected Calibration Error (ECE) \cite{naeini2015obtaining} is perhaps the most commonly used metric for estimating calibration discrepancy.
ECE is a discrete empirical approximation of the expected absolute difference between prediction accuracy and confidence estimation.
The ECE is formally given by:

\begin{equation}
\operatorname{acc}\left(B_{m}\right)=\frac{1}{\left|B_{m}\right|} \sum_{i \in B_{m}}, \mathds{1}\left(\hat{y}_{i}=y_{i}\right),
\label{eq:acc}
\end{equation}

\begin{equation}
\operatorname{conf}\left(B_{m}\right)=\frac{1}{\left|B_{m}\right|} \sum_{i \in B_{m}}, \hat{q}_{i},
\label{eq:conf}
\end{equation}

\begin{align}
    \operatorname{ECE} = \sum_{m=1}^B\frac{\left|B_{m}\right|}{N}|\acc(B_m) - \conf(B_m)|.
    \label{eq:ECE}
\end{align}
Here, $B_m$ denotes the set of samples belonging to the $\text{m}^{th}$ bin, $\left|B_{m}\right|$ is the number of instances residing in bin $b$, $N$ is the total number of samples, $B$ is the total number of bins and $\mathds{1}$ is the indicator function.

Since prediction accuracy cannot be estimated for individual samples but rather by taking the mean accuracy over a group of samples, the ECE score employs a binning scheme aggregating close by confidence values together. There is a resolution-accuracy trade-off between choosing more or fewer bins, and bin boundaries should be chosen carefully. During our experimentation, we choose an adaptive binning strategy proposed by \cite{Nixon_2019_CVPR_Workshops}, where the boundaries are set such that they split the samples into $B$ even groups of $N/B$ samples each. 

This scheme adapts the bins to the natural distribution of confidence scores, thus, trading-off resolution between densely and sparsely populated confidence regions while keeping the accuracy estimation error even among the bins. We refer our readers to \cite{on_calib_icml17} for details and experimentation with different variations of the ECE metric.

\paragraph{Negative log likelihood}
The Negative Log Likelihood (NLL) objective is commonly used for classifier confidence calibration. NLL is defined as:

\begin{equation} 
\mathcal{L}=-\sum_{i=1}^{n} \log \left(\hat{\pi}\left(y_{i} \mid \mathbf{x}_{i}\right)\right),
\label{eq:nll}
\end{equation}

where the estimated probability $\hat{\pi}$ for the ground truth label $y_i$ given the sample $x_i$ is formulated as

\begin{equation*}
\hat{\pi}\left(y_{i} \mid \rvx_i\right) = \softmax \left( \rvz_{i}\right)^{(y_i)}.
\label{eq:nll-prob}
\end{equation*}

\paragraph{Brier Score}
Brier score~\cite{brier1950verification} is a scoring method developed in an effort to predict the reliability of weather forecasts and has been subsequently adapted as a proxy for calibration error. Since the number of possible sequential labels is intractable, we treat the problem as a one vs. all classification, enabling the use of the binary Brier formulation. The Brier score as formalized in Equation \ref{eq:brier} is the mean square error between the confidence scores and the binary indicator function over the predicted and ground truth labels. 

\begin{align}
    \operatorname{Brier} = \sum_{i=1}^N(\mathbf{1}\left(\hat{y}_{i}=y_{i}\right) - \hat{q}_{i})^2.
    \label{eq:brier}
\end{align}

According to ~\cite{murphy1973new}, the Brier score comprises three components: uncertainty, reliability, and resolution. While confidence calibration is tasked to minimize the reliability term, the other terms carry information regarding the data uncertainty and deviation of the conditional probabilities from the mean.
Therefore, while the Brier score contains a calibration error term, it is entangled with two other terms leading to sub-optimal calibration. 

The main advantage of minimizing the Brier score is that it is parameter independent as it does not depend on data binning. in Section \ref{sec:experiments}, we demonstrate that minimization of Brier score leads to reduced ECE on a held-out test-set.

\section{Sequence-Level Calibration}
\label{sec:framework}

We propose to incorporate existing and new calibration methodologies while optimizing for word-level confidence scores, as depicted in Figure \ref{fig:optimzation}~(b). Our optimization scheme consists of a calibration model 
and a calibration objective applied to word-level scalar confidences.

Starting with a pre-trained STR model, we freeze the model weights and apply the T-scaling calibration method by multiplying the logits for each decoding step by the temperature parameter $T$. Following Kuleshov and Liang~\cite{Kuleshov2015} we define the ``Event of Interest'' as the exact word match between predicted and ground-truth words. 
For each word prediction, we define a scalar confidence score as the product of individual decoding-step confidences. We assess our performance according to the ECE metric with equal-sized bins as the work by Nixon~\etal~\cite{Nixon_2019_CVPR_Workshops} suggests.

\paragraph{Calibration of Non-IID Predictions}
We motivate our choice of word-level calibration from a probabilistic viewpoint. Taking into account inter-sequence dependencies, we assume that the predictions made at each decoding step are non-IID. This is especially evident for RNN-based decoders \eg Attention-based decoders, where each decoded character is provided as an input for the following prediction.
In this case, the following inequality holds:
\begin{equation}
    \mathbb{P}(\hat{Y}=Y|x) \neq \prod_i \mathbb{P}(\hat{y_i} = y_i | x).
    \label{eq:iid_decoding}
\end{equation}
Here, $\mathbb{P}(\hat{Y}=Y|x)$ denotes the correct prediction probability for the predicted sequence $\hat{Y}$ for input $x$, $\mathbb{P}(\hat{y_i}= y_i |x)$ are the marginal probabilities at each decoding step and $ \hat{Y}=(\hat{y_1}, ..., \hat{y_L}) $ are the predicted sequence tokens.

The calibration process attempts to affect the predicted scores so that they tend towards the prediction probabilities. Therefore, Equation \ref{eq:iid_decoding} implies that calibration of the marginals $\mathbb{P}(\hat{y_i} = y_i|x)$, corresponding to character-level calibration (Figure \ref{fig:optimzation}~(a)), will not lead to a calibrated word-level confidence (Figure \ref{fig:optimzation}~(b)).
This insight leads us to  advocate for direct optimization of the left hand side of Equation~\ref{eq:iid_decoding} \ie the word-level scalar confidence scores.

\paragraph{Objective Function}

In previous work, several calibration objective functions have been proposed. Three of the commonly used functions are ECE, Brier, and NLL.
Typically T-scaling is optimized via the NLL objective. Since the proposal of ECE by Naeini~\etal~\cite{naeini2015obtaining}, it has been widely adopted as a standard calibration measure. Adopting the findings of Nixon~\etal~\cite{Nixon_2019_CVPR_Workshops}, we utilize ECE as the calibration optimization objective. In our experiments, we also examine the Brier and NLL objectives. We find that while Brier can reduce ECE, it does not converge to the same minima. As for the NLL score, we find that it is unsuitable for sequence-level optimization as directly applying it to multiplied character scores achieves the same minima as character-level optimization. This is undesirable due to the inequality from Equation \ref{eq:iid_decoding}. (see supplementary for more details).

\paragraph{Step Dependent T-Scaling (STS)}
\label{sec:calib_model_timestamp}
We extend T-Scaling to better suit the case of sequence prediction. As T-Scaling applies a single, global parameter to all model outputs, it does not leverage existing inter-sequence dependencies. This is especially true for context-dependent models such as Attention and Transformer based decoders. Therefore, we propose extending the scalar T-scaling to  \emph{Step Dependent T-scaling (STS)} by setting an individual temperature parameter for each character position within the predicted sequence.
We replace the scalar temperature $T$ with a vector $\mathbf{T} \in \R^{\tau + 1}$, $\mathbf{T} = \{T_0, ..., T_{\tau}\}$, where $\tau$ is a truncation length.
This may be formulated as:
\begin{equation}
    \hat{q}_{i,j} = \max_k \softmax \left( \rvz_{i,j} \mathbf{T}_j \right)^{(k)}.
    \label{eq:time_temp_scaling}
\end{equation}
Here, $\rvz_{i,j}$ is the logits vector for the $\text{j}^{th}$ character of the $\text{i}^{th}$ sample, and $\mathbf{T}_j$ is the temperature value applied to the $\text{j}^{th}$ character for all sequences.

Applying this method directly, however, results in sub-optimal results. This is due to the increase in trainable parameters for the same size calibration set. Furthermore, longer words are scarce and present high variability; thus, it may skew the temperature values of time steps above a certain index.
We propose a meta-parameter $\tau$ that applies to all time steps over a certain value, such that $\mathbf{T}_{j_\geq \tau} = \mathbf{T}_\tau$.

\paragraph{Edit Distance Expected Calibration Error (ED-ECE)}
\label{sec:framework_ed_calib}
A single classification is either correct or incorrect in its prediction. Sequential predictors, however, present a more nuanced sense of correct prediction \eg correctly predicting $4$ out of $5$ characters is not as bad as predicting only $3$.
When calibrating in order to minimize the accuracy-confidence gap, the estimated accuracy is obtained by Equation \ref{eq:acc}.
The indicator function implies a binary classification task, or in the multi-label setting, a one versus all classification. Sequence prediction, however, is more nuanced and allows for partial errors. We propose to incorporate the error rate into a new calibration metric. By doing so we allow the end user to assess not only the probability for absolute error but also the amount of incorrect predictions within the sequence. To this aim, we propose to manipulate the ECE metric from by replacing Equation \ref{eq:acc} with the following:

\begin{equation}
\operatorname{acc_{n}}\left(B_{m}\right)=\frac{1}{\left|B_{m}\right|} \sum_{i \in B_{m}} \mathbf{1}\left(\operatorname{ED}(\hat{y}_{i},y_{i}) \leq n \right).
\label{eq:ed_calibration}
\end{equation}

Here, $\operatorname{ED}$ refers to the Edit Distance function, also know as the Levenshtein Distance \cite{levenshtein1966binary}. The Edit Distance produces an integer enumerating the minimal number of insertions, deletions, and substitutions performed on one string to produce the other. 

We term our modified ECE metric -- Edit Distance Expected Calibration Error (ED-ECE). Essentially, ED-ECE is a fine-grained per-character-centric metric, whereas ECE is a coarse-grained word-centric metric. ED-ECE can be minimized for any desired string distance $n$ and produces a confidence scores signifying the likelihood of an erroneous prediction up to an Edit Distance of $n$. This information is helpful for downstream applications such as dictionary lookup, beam-search or human correction, as each example may be sent down a different correction pathway according to some decision scheme.
\section{Experiments}
\label{sec:experiments}

In the following section, we carry out an extensive evaluation and analysis of our proposed optimization framework. We begin by detailing our experimental setup, including datasets, evaluated models, optimization methodology, and implementation details
in~Section~\ref{sec:exp_setup}.

In Section~\ref{sec:results_analysis} we provide a deeper analysis, including sequence calibration for various aggregation window sizes (n-grams lengths). We further provide detailed results and analysis for the aspects described thus far: ED-ECE metric, calibration by decoding T-scaling step (STS), and the gains obtained through calibration and beam-search based decoding.

\subsection{Experimental Setup} \label{sec:exp_setup}
\paragraph{Datasets}

For the sake of fairness, all evaluated STR models are retrained using the same data, specifically the SynthText~\cite{Zisserman206st} dataset. Models are then evaluated using both regular and irregular text datasets.
Regular text datasets containing text with nearly horizontally aligned characters include: IIIT5K~\cite{Mishra2012sj}, SVT~\cite{Wang2011bottom}, ICDAR2003~\cite{Lucas2003ic03}, and ICDAR2013~\cite{Karatzas2013ic13}.
Irregular text datasets are comprised of arbitrarily shaped text (\eg curved text), and include include: CDAR2015~\cite{Karatzas2015ic15}, SVTP~\cite{Phan2013svtp}, and CUTE 80~\cite{Risnumawan2014cute}.

\paragraph{Text Recognition Models}
\label{sec:exp-models}

Our experiments focus on several recent STR models.
Baek~\etal~\cite{Baek2019clova} proposed a framework text recognition model comprising four stages: transformation, feature extraction, sequence modeling, and decoding. We consider architectures using four of their proposed variants, including or omitting BiLSTM combined with either a CTC~\cite{Graves2006ctc} or an attention~\cite{Chen2016rnnocr} decoder.
In ASTER~\cite{Bai2018aster}, oriented or curved text is rectified using an STN~\cite{jaderberg2015spatial}, and a BiLSTM is used for encoding. A GRU~\cite{cho-etal-2014-learning} with an attention mechanism is used for decoding.
SCATTER~\cite{Litman_2020_CVPR} uses a stacked block architecture that combines both visual and contextual features.
They also used a two-step attention decoding mechanism for providing predictions.
Bartz~\etal~\cite{bartz2019kiss} proposed KISS, combining region of interest prediction and transformer-based recognition.
SEED~\cite{qiao2020seed} is an attention-based architecture supplemented with a pre-trained language model.

In our work, we evaluate and analyze each of these models' calibration-related behavior, highlighting differences between various decoder types.

\paragraph{Optimization}
The task of calibrating confidence scores boils down to minimizing the model parameters w.r.t a given loss function. T-scaling based calibration methods take the predicted logits $\rvz_i$ as input and apply a modified SoftMax operation to arrive at the calibrated confidence score.

Backpropagation is only conducted through the optimized calibration parameter, while the STR model remains unchanged. The model formulations for T-scaling and STS are provided in Equation~\ref{eq:temp_scaling} and Equation~\ref{eq:time_temp_scaling} respectively.

We use the L-BFGS optimizer ~\cite{liu1989limited} coupled with several calibration objective functions to demonstrate our calibration methodology. Our tested loss functions include ECE, ED-ECE, Brier, and NLL. Table~\ref{tab:loss_compare} presents ECE achieved by calibrating for NLL, Brier, and the ECE objective functions. We observe that ECE obtains the best calibration error as expected while Brier succeeds to a lower degree. We also demonstrate that, as mentioned, NLL is not suitable for word-level calibration.

\begin{table}[t!]
	\centering
	\setlength{\tabcolsep}{4pt}
\begin{tabular}{l cc cc} %
\toprule

\textbf{Method} &      Uncalib. & ECE & Brier & NLL \\
\midrule	
CTC~\cite{Baek2019clova}           & 6.9     & 2.2            & 5.9          & 6.8 \\ 
BiLSTM CTC~\cite{Baek2019clova}    & 6.7     & 2.0            & 6.0          & 6.6 \\ 
Atten.~\cite{Baek2019clova}        & 5.9     & 1.8            & 5.2          & 5.9 \\ 
BiLSTM Atten.~\cite{Baek2019clova} & 5.4     & 2.0            & 4.8          & 5.3 \\ 
ASTER~\cite{Bai2018aster}          & 1.8     & 0.8            & 1.8          & 1.7 \\ 
SCATTER~\cite{Litman_2020_CVPR}    & 5.8     & 1.8            & 4.5          & 5.7 \\ 
KISS~\cite{bartz2019kiss}          & 9.6     & 1.4            & 5.3          & 9.4 \\ 
SEED~\cite{qiao2020seed}           & 5.7     & 2.0            & 5.7          & 5.6 \\ 
\midrule
Average                            &5.98     & \textbf{1.75}  & 4.9          & 5.88 \\ 
\bottomrule
\end{tabular}

    \caption{\textbf{Calibrated ECE Scores for Different Objective Functions.} Unsurprisingly, ECE values are best optimized w.r.t. the ECE objective. Brier loss is also suitable for reducing calibration error but is less effective. Finally, we observe that NLL is unsuitable for sequence level calibration as detailed in Section~\ref{sec:framework} and in the supplementary.}
	\label{tab:loss_compare}
\end{table}

\begin{figure}[h]
\begin{center}
  \includegraphics[width=\linewidth]{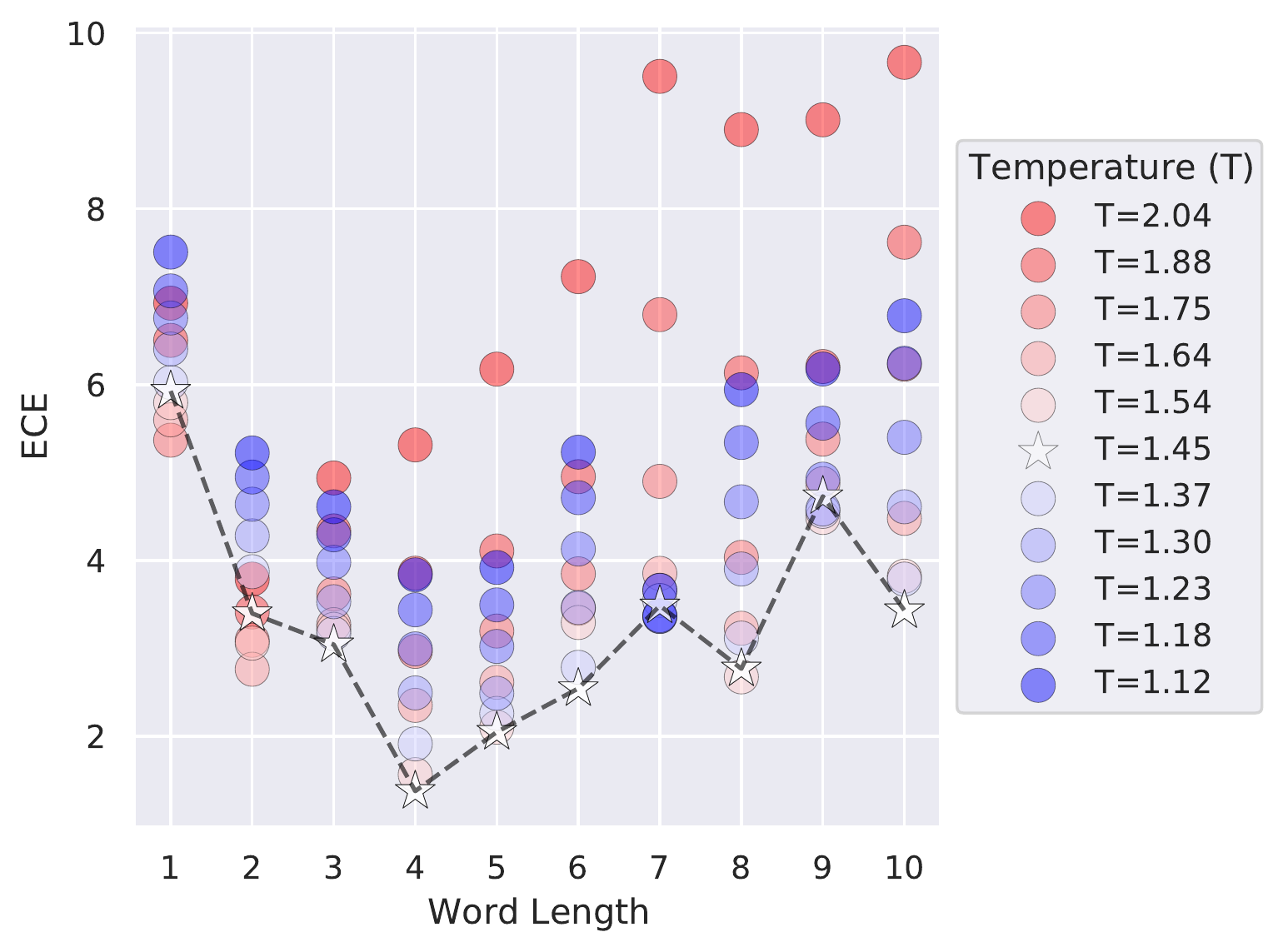}
\vspace{-0.6cm}
\end{center}
   \caption{\textbf{ECE for Different Word Lengths and Temperatures}. Here, $T=1.45$ is the optimal T-scaling temperature marked with a white star, warmer temperatures are marked with red circles and colder ones in blue. We observe that the optimal temperature for the AttDec STR model is globally optimal for almost all word lengths. For word lengths of one or two characters a sightly higher temperature would achieve marginal improvement in ECE. See supplementary for additional models.}
   \label{fig:calib_by_length}
\end{figure}
\vspace{-0.3cm}

\subsection{Results and Analysis} \label{sec:results_analysis}

\paragraph{Aggregation Window} \label{sec:exp_agg-window}
In order to gain a deeper understanding of the relation between aggregation and calibration performance, we experimented with calibration via partial sequences. To this end, we break up our calibration datasets into all possible sub-sequences of $\text{length} \leq n$. We note that when $n=1$ the calibration is carried out at character-level as depicted by Figure~\ref{fig:optimzation}~(a), and for $n= \max \length (w_i)$ we are calibrating on the full sequence (Figure~\ref{fig:optimzation} ~(b) )

\begin{figure*}[h]
\begin{center}
  \includegraphics[width=0.40\textwidth]{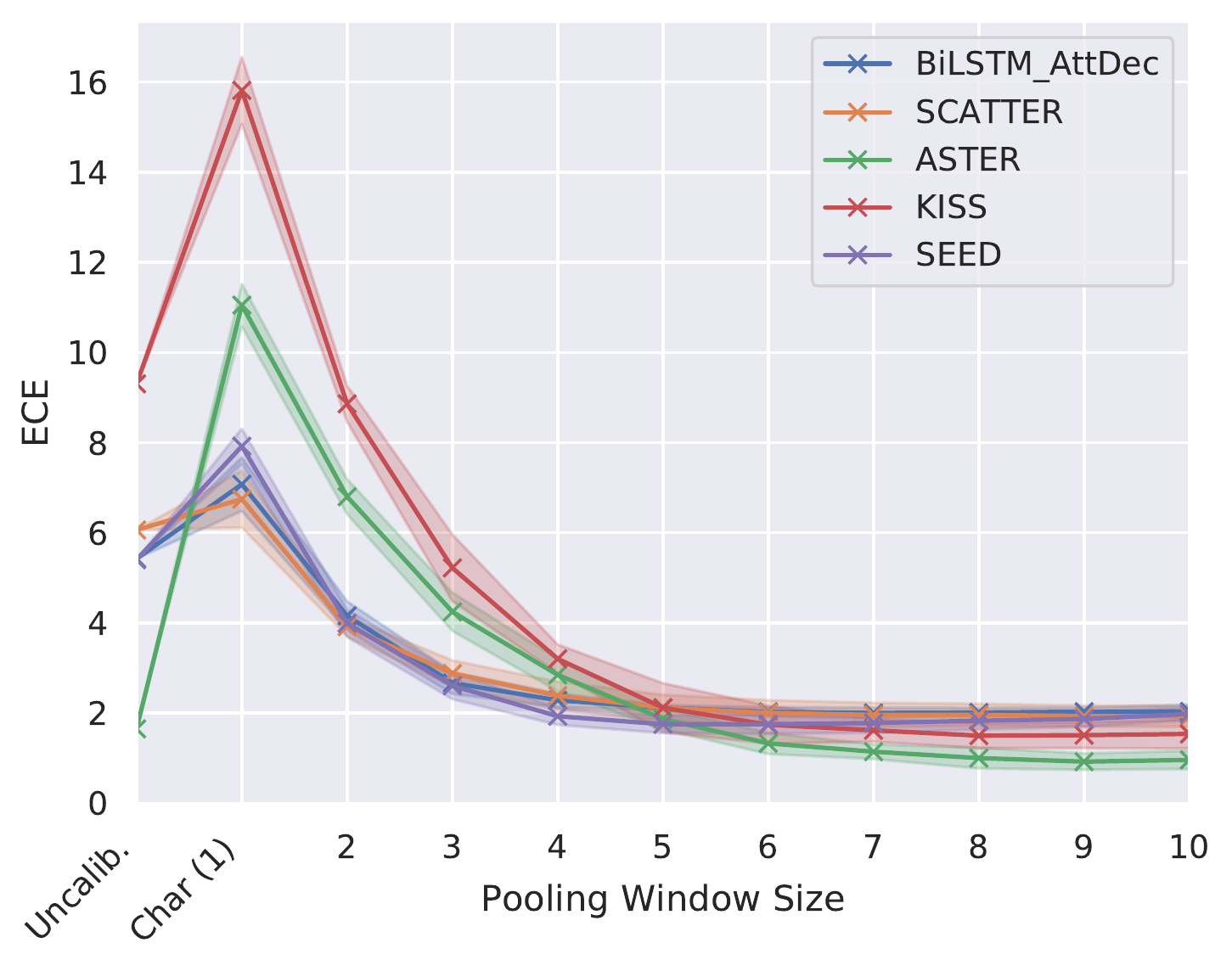}
  \hspace{0.02 \textwidth}
  \includegraphics[width=0.40\textwidth]{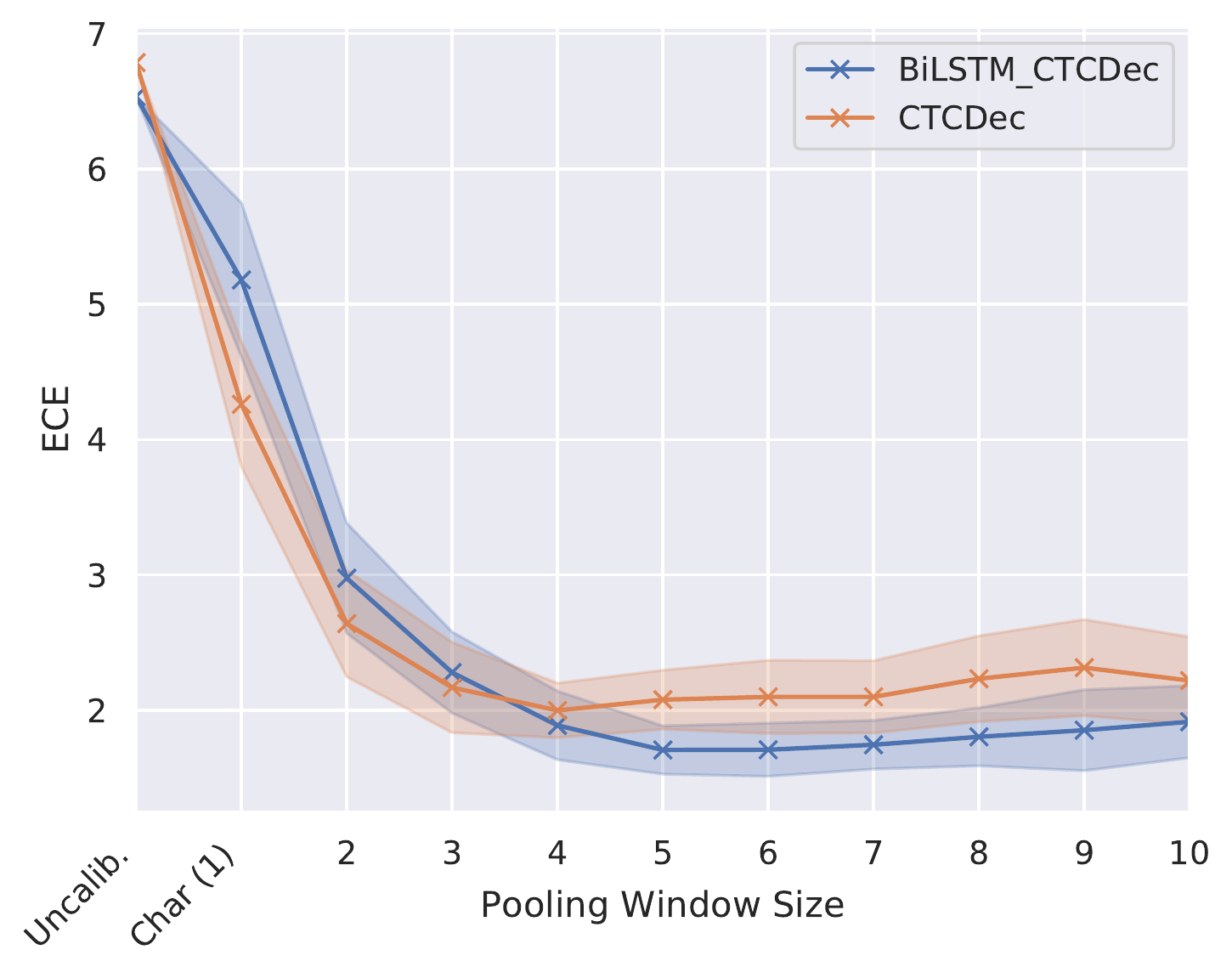}
\vspace{-0.5cm}
     \caption{\textbf{Calibrated Word-Level ECE Values vs. Pooling Window Sizes (n-grams)} for different STR methods.
     We evaluate our results on $10$ saved checkpoints for each model, plotting the mean and standard deviation.
     \textbf{Left:} Attention based decoders. \textbf{Right:} CTC based decoders. All results are demonstrated on a held-out test-set. ``Char'' refers to character level calibration corresponding to a window size of $1$. We observe that attention based models become uncalibrated when individual character calibration is performed and that longer window sizes are preferred over short ones during calibration.}
   \label{fig:ngrams_vs_ece}
   \end{center}
\vspace{-0.5cm}
\end{figure*}

Calibration is performed by the T-scaling method coupled with the ECE objective function. All reported results are measured on a held-out test-set. In an attempt to reduce noise, we test the calibration process on $10$ training checkpoints of each model and plot the mean and variance measurements in Figure~\ref{fig:ngrams_vs_ece}.

We find that for attention decoders (Figure~\ref{fig:ngrams_vs_ece} ~(Left) $n=1$ provides worse calibration than the uncalibrated baseline. CTC decoders (Figure~\ref{fig:ngrams_vs_ece} ~(Right), on the other hand, also exhibit worse ECE scores on per-character calibration; however, the error is still reduced relative to the uncalibrated models. We postulate that this phenomenon relates to the difference between IID and non-IID decoding discussed in Section~\ref{sec:framework}. This key observation emphasizes the importance of score aggregation during the calibration process as opposed to individual character calibration.

\begin{table}[t]
	\centering
	\setlength{\tabcolsep}{8pt}
\begin{tabular}{l cc cc} %
\toprule

 & \multicolumn{2}{c}{$ \operatorname{E_d} \: \leq \: 1$} & \multicolumn{2}{c}{$ \operatorname{E_d} \: \leq \: 2$} \\ 
\cmidrule(lr){2-3} \cmidrule(lr){4-5}
  { \textbf{Method} \textbackslash \textbf{Calibration}}       & X     & \large{$\checkmark$}       & X    & \large{$\checkmark$}       \\ 

\midrule	
CTC~\cite{Baek2019clova}           & 4.1    & 1.8  & 8.2  & 1.7  \\
BiLSTM CTC~\cite{Baek2019clova}    & 4.2    & 1.5  & 7.9  & 1.9  \\
Atten.~\cite{Baek2019clova}        & 2.0    & 0.9  & 4.5  & 0.7  \\
BiLSTM Atten.~\cite{Baek2019clova} & 2.1    & 1.0  & 5.0  & 0.7  \\
ASTER~\cite{Bai2018aster}          & 5.3    & 2.1  & 9.7  & 1.6  \\
SCATTER~\cite{Litman_2020_CVPR}    & 1.6    & 1.2  & 3.9  & 0.9  \\
KISS~\cite{bartz2019kiss}          & 1.0    & 0.8  & 5.2  & 1.2  \\
SEED~\cite{qiao2020seed}           & 2.1    & 1.2  & 4.9  & 0.7  \\
\midrule
Average                            & 2.8 & 1.31 & 6.16 & 1.18   \\ 
\bottomrule
\end{tabular}

  \caption{\textbf{ED-ECE Values} for uncalibrated (X) and calibrated ($\checkmark$) models. Calibration was performed by the T-scaling method and ED-ECE objective for $\operatorname{E_d} \leq 1,2$. We observe that the optimization process reduces ED-ECE values on the held-out test-set. The calibrated ED-ECE scores may be used to estimate the number of incorrect predictions within the sequence for down-stream applications.}
	\label{tab:framework_ED-ECE}
\end{table}

\paragraph{Edit Distance Expected Calibration Error} \label{sec:exp_ed-ece}
In Section~\ref{sec:framework_ed_calib} we present a new calibration metric termed Edit Distance Expected Calibration Error (ED-ECE). We calibrate for ED-ECE with $n=[1,2]$ and present the results in Table~\ref{tab:framework_ED-ECE}. As expected, the ED-ECE is reduced significantly due to the optimization. It is worth noting that ED-ECE is often lower than the original ECE score, leading to a more accurate confidence estimation. Once calibrated, three scores corresponding to ECE (ED-ECE for $n=0$) and ED-ECE for $n=[1,2]$ are produced for each data sample. This allows us to submit the data to further review according to thresholds on the output scores.

For example given a predicted label ``COFEEE'' for a ground-truth label of ``COFFEE'', the absolute and $\operatorname{E_d} = 1$ predicted confidences are $75\%$ and $98\%$ respectively. We might recognize such an example and perform a focused search by testing the confidence scores of words that differ by an Edit-Distance of $1$ and selecting the most confident prediction within the search space.

\begin{table}[t]
	\centering
    
\setlength{\tabcolsep}{8pt}
\begin{tabular}{l ccc} %
\toprule
  {\textbf{Method}}                & Uncalib.  & TS            & STS    \\ 
\midrule	
CTC~\cite{Baek2019clova}           & 6.9       &\textbf{2.2}   & \textbf{2.2}   \\
BiLSTM CTC~\cite{Baek2019clova}    & 6.7       & 2.0           & \textbf{1.8}   \\ 
Atten.~\cite{Baek2019clova}        & 5.9       & 1.8           & \textbf{1.7}   \\
BiLSTM Atten.~\cite{Baek2019clova} & 5.4       & 2.0           & \textbf{1.7}   \\
ASTER~\cite{Bai2018aster}          & 1.8       & \textbf{0.8}  & 1.0            \\
SCATTER~\cite{Litman_2020_CVPR}    & 5.8       & 1.8           & \textbf{1.6}   \\
KISS~\cite{bartz2019kiss}          & 9.6       & \textbf{1.4}  & \textbf{1.4}   \\
SEED~\cite{qiao2020seed}           & 5.7       & \textbf{2.0}  & \textbf{2.0}   \\
\midrule
Average                              & 5.98 & 1.75 & \textbf{1.67}    \\ 
\bottomrule
\end{tabular}

    \caption{\textbf{ECE Values Comparing T-scaling and STS} for uncalibrated (Uncalib.) and calibrated confidence scores obtained on a held-out test-set. We optimize according to our proposed method utilizing the T-scaling (TS) and the proposed STS calibration methods. We demonstrate that STS is slightly advantageous over global TS.}
    \label{tab:framework_ECE}
\end{table}
\begin{table*}[h]
	\centering

\setlength{\tabcolsep}{7pt}
\begin{tabular}{l c cc cc cc cc} %
\toprule

& bw=1& \multicolumn{2}{c}{bw=2} & \multicolumn{2}{c}{bw=3} & \multicolumn{2}{c}{bw=4} & \multicolumn{2}{c}{bw=5} \\  
   \cmidrule(lr){2-2} \cmidrule(lr){3-4} \cmidrule(lr){5-6} \cmidrule(lr){7-8} \cmidrule(lr){9-10}
  {\textbf{Method}}\textbackslash{\textbf{Calibration}}   &  X  & X   & \large{$\checkmark$}   & X      &  \large{$\checkmark$}    & X     &  \large{$\checkmark$}     & X    &  \large{$\checkmark$}           \\ 

\midrule	
Atten.~\cite{Baek2019clova}          & 86.21 &  +0.18 &	+0.39 &	+0.23 &  +0.43    &	+0.26 & +\bf{0.47} &	+0.25 & +0.45      \\
BiLSTM Atten.~\cite{Baek2019clova}   & 86.2  &  +0.11 &	+0.2  &	+0.18 &	+0.33    &	+0.2  & +\bf{0.35} &	+0.21 & +\bf{0.35} \\
ASTER~\cite{Bai2018aster}            & 86.03 &	+0.52 &	+0.57 &	+0.63 &  +0.77    &	+0.64 & +\bf{0.82} &	+0.64 & +0.81      \\
SCATTER~\cite{Litman_2020_CVPR}      & 87.36 &  +0.16 &	+0.27 &	+0.2  & +\bf{0.3} &	+0.19 &	   +0.28  &	+0.2  & +0.29      \\
SEED~\cite{qiao2020seed}             & 81.18 &	+0.2  &	+0.29 &	+0.23 &  +0.35    &	+0.32 & +\bf{0.42} &	+0.28 & +0.4       \\
\midrule
Average                              & 84.53 &	+0.23 &	+0.34 &	+0.29 &	+0.44    &	+0.32 & +\bf{0.47} &	+0.32 & +0.46      \\ 
\bottomrule
\end{tabular}

    \caption{\textbf{Beam-search accuracy gains} achieved for calibrated ($\checkmark$ ) vs.  uncalibrated (X) models. We apply beam widths (bw) between $1$ and $5$. Displayed results are averaged across test datasets and are reported relative to the baseline of $\text{bw}=1$, which is equivalent to not using beam-search. CTC based methods are omitted as beam-search requires decoding dependence in order to be effective. We demonstrate consistent improvement across all models and all datasets (see supplementary for breakdown) over the uncalibrated baseline.}
    \label{tab:beam-search}
	\vspace{-1ex}
\end{table*}
\paragraph{Calibration by Word Length}
\label{sec:exp_calib_by_length}
In this experiment we explore the calibration characteristics of words with varying word lengths. We calibrate an Attention decoder variant of~\cite{Baek2019clova} using T-Scaling for STR and evaluate the ECE for different word lengths. The model calibration produces an optimal value of $T=1.45$ for the calibration temperature.
We perturb the temperature around the optimal value and plot the results in Figure~\ref{fig:calib_by_length}. As can be observed in the figure, the temperature resulted from sequence-level calibration is optimal for almost all word lengths. This finding is surprising since one could expect the calibration to be optimal for some word lengths and not others. Another observation we derive from Figure~\ref{fig:calib_by_length} is that single characters are the least calibrated, where 4 and 5 letter words are the most frequent in the English language (and in the test dataset) achieve the best calibration among all groups.

\paragraph{Step Dependent T-Scaling (STS)}
In Section~\ref{sec:calib_model_timestamp} we propose \emph{Step Dependent T-Scaling (STS)}. STS extends the previously presented T-scaling by assigning a temperature for each character position in the sequence. Table~\ref{tab:framework_ECE} lists the calibrated ECE values achieved by T-scaling as well as STS calibration schemes coupled with an ECE calibration objective function. In all experiments, we select $\tau=5$ temperature values while a \nth{6} value is used to calibrate the subsequent sequence positions. Our experimentation demonstrates that the Time-Stamp scaling is beneficial or on par with T-scaling for all but one of the models. Overall, when averaging on all tested models, STS shows a slight benefit over T-scaling. We hypothesize that STS is able to improve calibration error due to its finer-grained calibration and exploitation of inter-sequence relations.

\paragraph{Beam-Search} \label{sec:exp_beam_search}
Although calibration methods based on T-scaling do not alter prediction accuracy, it is still possible to indirectly affect a model's accuracy rate. This can be achieved through a beam-search methodology, where the space of possible predicted sequences is explored within a tree of possible outcomes. At each leaf, the total score is calculated as the product of all nodes leading up to the leaf.

The dependence of each decoded character on previous predictions means that, although the confidence ordering is preserved among possible outcomes at each step, the scores do change in such a way that reorders the aggregated scores within the search. However, this is not true for stateless decoders such as CTC, which are therefore omitted from this experiment.

We present our calibrated beam-search results in Table~\ref{tab:beam-search}, showing a consistent gain for each calibrated mode1 relative to the non-calibrated baseline. We also show that this holds for all tested beam widths between two and five. In the supplementary material, we further break down the results according to the individual test datasets.

\section{Conclusion} %
In this work, we demonstrate that word-level and, in general, sequence-level calibration should be optimized directly on the per-word scalar confidence outputs. This is motivated by probabilistic reasoning and demonstrated empirically for various STR methods.

To the best of our knowledge, we are the first to conduct an in-depth analysis of the current state of calibration in scene-text recognition models. We perform extensive experimentation with STR model calibration and extend the T-scaling calibration method to a sequence-level variant we termed Step dependent T-Scaling (STS). We analyze the robustness of T-scaling to word length variation, finding that the T-scaling method achieves a calibrated temperature optimal for almost all word-lengths.

We also propose ED-ECE, a text-oriented metric and loss function, extending ECE to calibrate for sequence-specific accuracy measures (\eg Edit-Distance). We empirically show that ED-ECE achieves lower calibration error-rates than ECE.

Finally, we demonstrate that the calibration of STR models boosts beam-search performance, consistently improving model accuracy for all beam-widths and datasets.

\pagebreak

{\small
\bibliographystyle{ieee_fullname}
\bibliography{refs.bib}
}

\clearpage
\appendix

\section{Introduction}

In Section \ref{sec:implementation_supp} we present additional implementation detail regarding our training and re-implementation of the methods discussed in the paper (\cite{Baek2019clova,Litman_2020_CVPR,Bai2018aster,bartz2019kiss,qiao2020seed}). In Section \ref{sec:reliability_diagrams_supp} we present reliability plots for all models. Section \ref{sec:nll_supp} presents an extended discussion regarding the use of NLL loss in the context of sequence prediction models. Section \ref{sec:calib_word_length_supp} extends the calibration per-word length diagram by presenting diagrams for five of the models. Finally, Section \ref{sec:beam-search_supp} presents extended beam-search results for beam-width of four with per-dataset resolution.
\section{Implementation Details}  
\label{sec:implementation_supp}

To conduct a fair analysis, we have re-trained the STR models using the same data. Several methods such as \cite{bai2016tps},\cite{Baek2019clova} and \cite{bartz2019kiss} provide useful code. The original code was used with minor modifications, including the setting the model character-set and training dataset. The authors of \cite{Litman_2020_CVPR} and \cite{qiao2020seed} did not supply code, and we extended the code provided by Baek \etal \cite{Baek2019clova} according to the details in their papers. All models were trained solely on the SynthText dataset ~\cite{Zisserman206st} with a capital insensitive alphanumeric character-set. In addition, all non-alphanumeric symbols were mapped to an \emph{``UNKNOWN''} token. Achieved accuracy for the methods broken-down by dataset are presented in Table \ref{tab:reproduced_results}. We note that all results differ from the original published papers due to our re-implementation and reduced training dataset.

\begin{table*}[h]
	\centering
    
\begin{tabular}{l c c c c c c c c c c}
\toprule
\multirow{2}{*}{Method} & \multirow{2}{*}{BLSTM} & \multirow{2}{*}{DEC} &  
                                                                                                                     \begin{tabular}[c]{@{}c@{}}CUTE\\ 288\end{tabular} & \begin{tabular}[c]{@{}c@{}}IC03\\ 867\end{tabular} & \begin{tabular}[c]{@{}c@{}}IC13\\ 1015\end{tabular} & \begin{tabular}[c]{@{}c@{}}IC15\\ 2077\end{tabular} & \begin{tabular}[c]{@{}c@{}}IIIT5k\\ 3000\end{tabular} & \begin{tabular}[c]{@{}c@{}}SVT\\ 647\end{tabular} & \begin{tabular}[c]{@{}c@{}}SVTP\\ 645\end{tabular} & Avg \\ \midrule
CTC~\cite{Baek2019clova}    & No                     & CTC                  & 71.2     & 90.9     & 89.3      & 69.6      & 89.3        & 81    & 71.2     & 82.1 \\ 
BiLSTM CTC~\cite{Baek2019clova}      & Yes                    & CTC                  & 75.7     & 90.3     & 89.6      & 70.7      & 89.3        & 81.5    & 73.5     & 82.7 \\ 
Atten.~\cite{Baek2019clova}     & No                     & Atten.               & 77.4     & 93.3     & 92.5      & 75.4      & 92.2        & 85.5    & 78.5     & 86.2 \\ 
BiLSTM Atten.~\cite{Baek2019clova}       & Yes                    & Atten.               & 78.5     & 93.1     & 91.9      & 75.1      & 92        & 86.9    & 79.4     & 86.2 \\ 
ASTER~\cite{Bai2018aster}                    & Yes                    & Atten.               & 76.4     & 90.7     & 87.4      & 67.8      & 87.6        & 82.3    & 72.9     & 81.2 \\ 
SCATTER~\cite{Litman_2020_CVPR}                 & Yes                    & Atten.               & 83.7     & 93.8     & 93.1      & 77.4      & 92.6        & 87.5    & 78.9     & 87.4 \\ 
KISS~\cite{bartz2019kiss}                     & No                     & Trans.               & 75       & 87.1     & 85.2      & 65        & 82.2        & 75      & 63.4     & 76.7 \\ 
SEED~\cite{qiao2020seed}                     & Yes                    & Atten.               & 77.1     & 92.6     & 90.4      & 75.3      & 92.5        & 86.4    & 78.8     & 86 \\ 
\bottomrule
\end{tabular}

    \caption{\textbf{Reproduced Model Results.} We carry out our experimentation on eight recognition models retrained on  SynthText (ST) \cite{Zisserman206st}. The models were trained based on original code or our re-implementation. Evaluation is performed on seven common scene-text datasets. This table provides the test-set accuracy attained by our trained models. We note that the presented results differ from original reports due to modification of training-set and decoder character-set.}
    \label{tab:reproduced_results}
\end{table*}
The code for KISS \cite{bartz2019kiss} can be found here: \\ \url{https://github.com/Bartzi/kiss}\\
The code for ASTER \cite{Bai2018aster} can be found here: \\ \url{https://github.com/bgshih/aster}\\
The code for \cite{Baek2019clova} can be found here: \\ \url{https://github.com/clovaai/deep-text-recognition-benchmark}

\section{Reliability Diagrams}
\label{sec:reliability_diagrams_supp}
\begin{figure*}[t!]
\begin{center}
    \vspace{-4pt}
  \includegraphics[width=0.26\textwidth]{latex/figures/fig1_AttDec.pdf}
  \hspace{-12pt}
  \includegraphics[width=0.26\textwidth]{latex/figures/fig1_CTCDec.pdf}
    \hspace{-12pt}
  \includegraphics[width=0.26\textwidth]{latex/figures/fig1_SCATTER.pdf}
    \hspace{-12pt}
  \includegraphics[width=0.26 \textwidth]{latex/figures/fig1_SEED.pdf}
  \hspace{-12pt}
  \includegraphics[width=0.26 \textwidth]{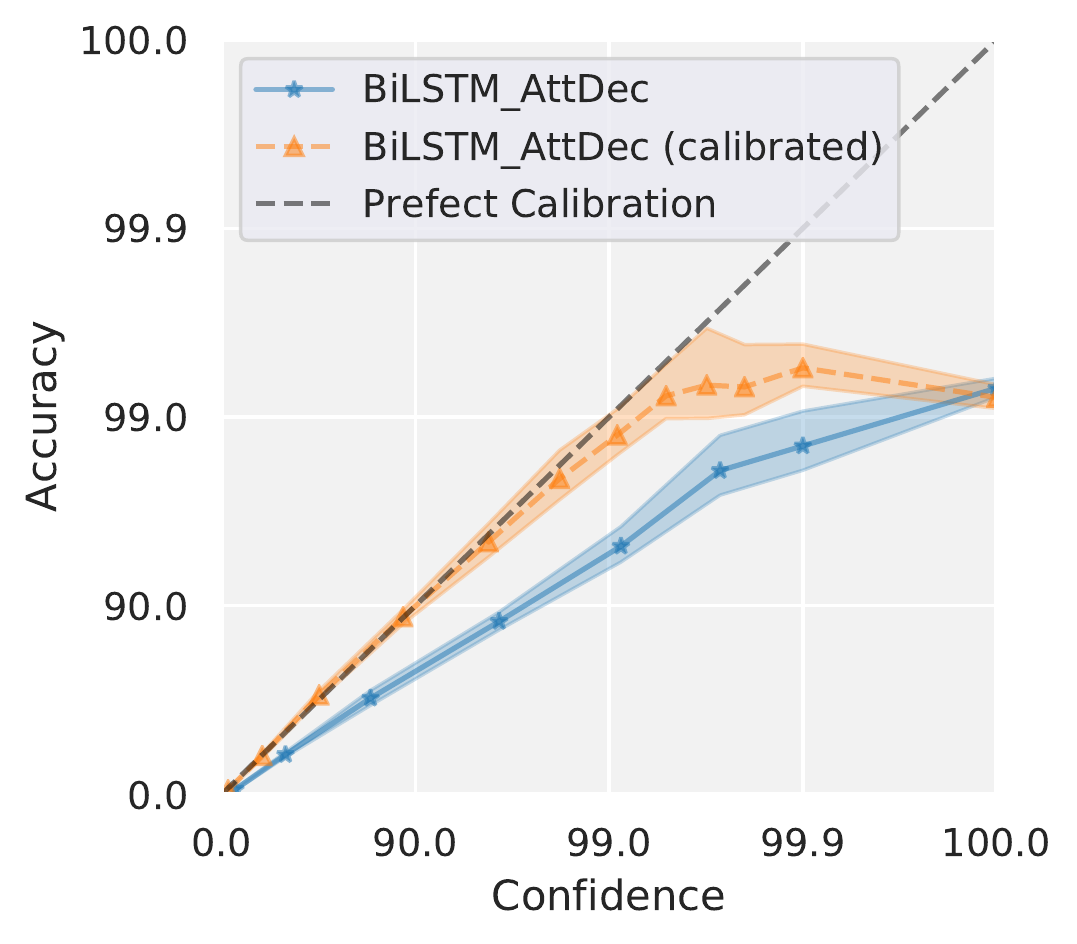}
  \hspace{-12pt}
  \includegraphics[width=0.26 \textwidth]{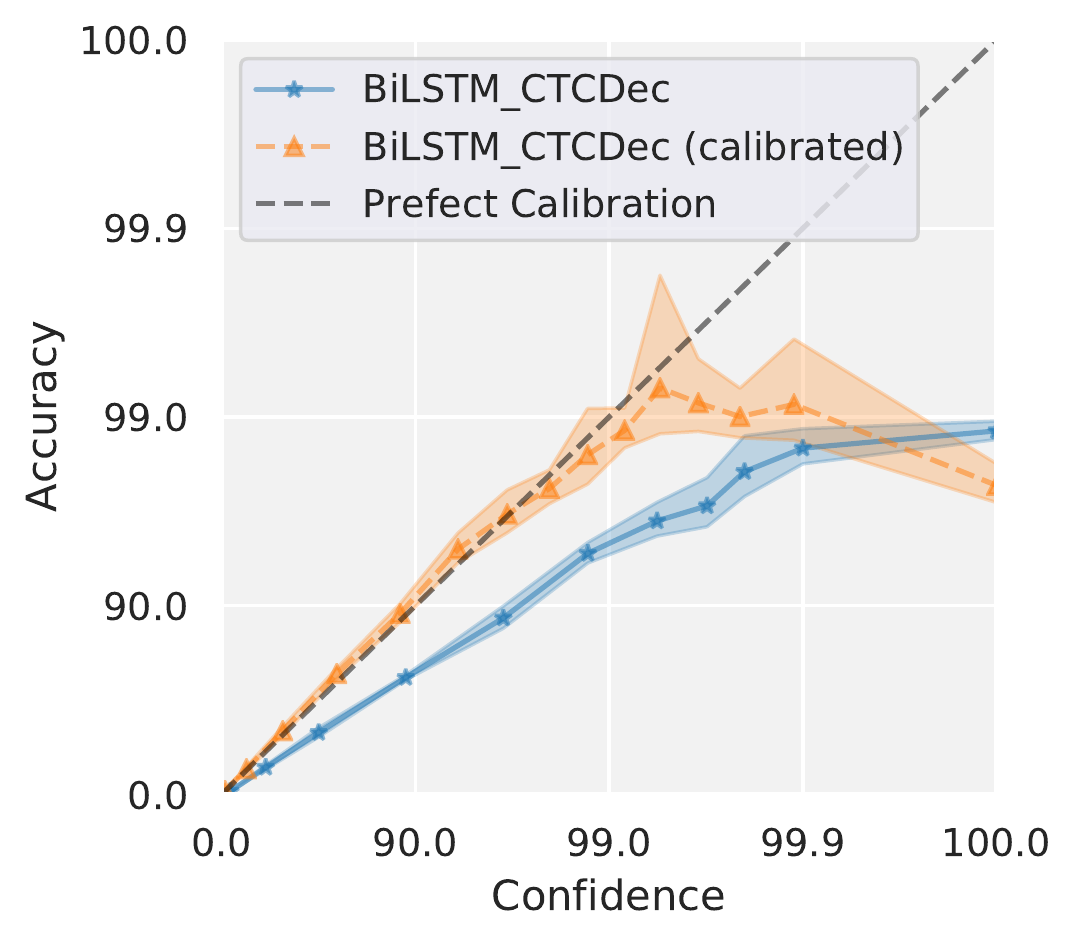}
  \hspace{-12pt}
  \includegraphics[width=0.26 \textwidth]{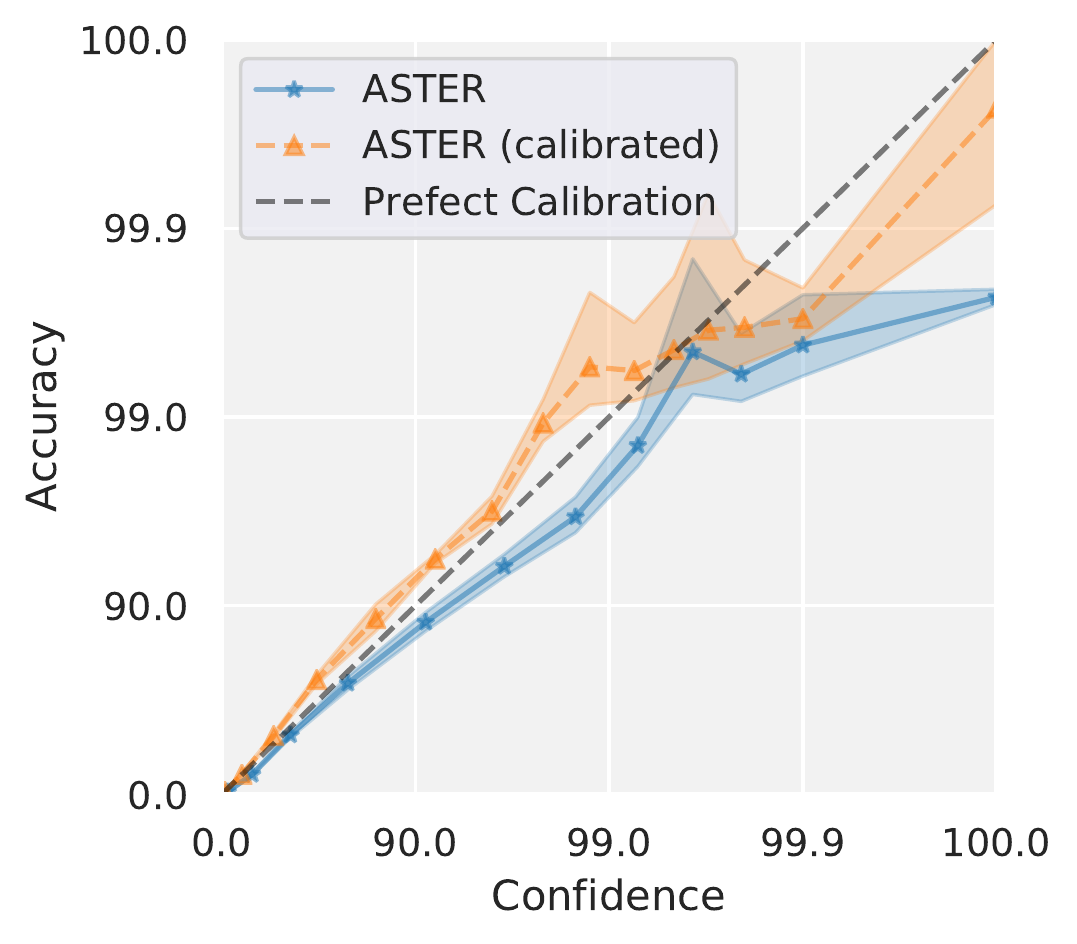}
  \hspace{-12pt}
  \includegraphics[width=0.26 \textwidth]{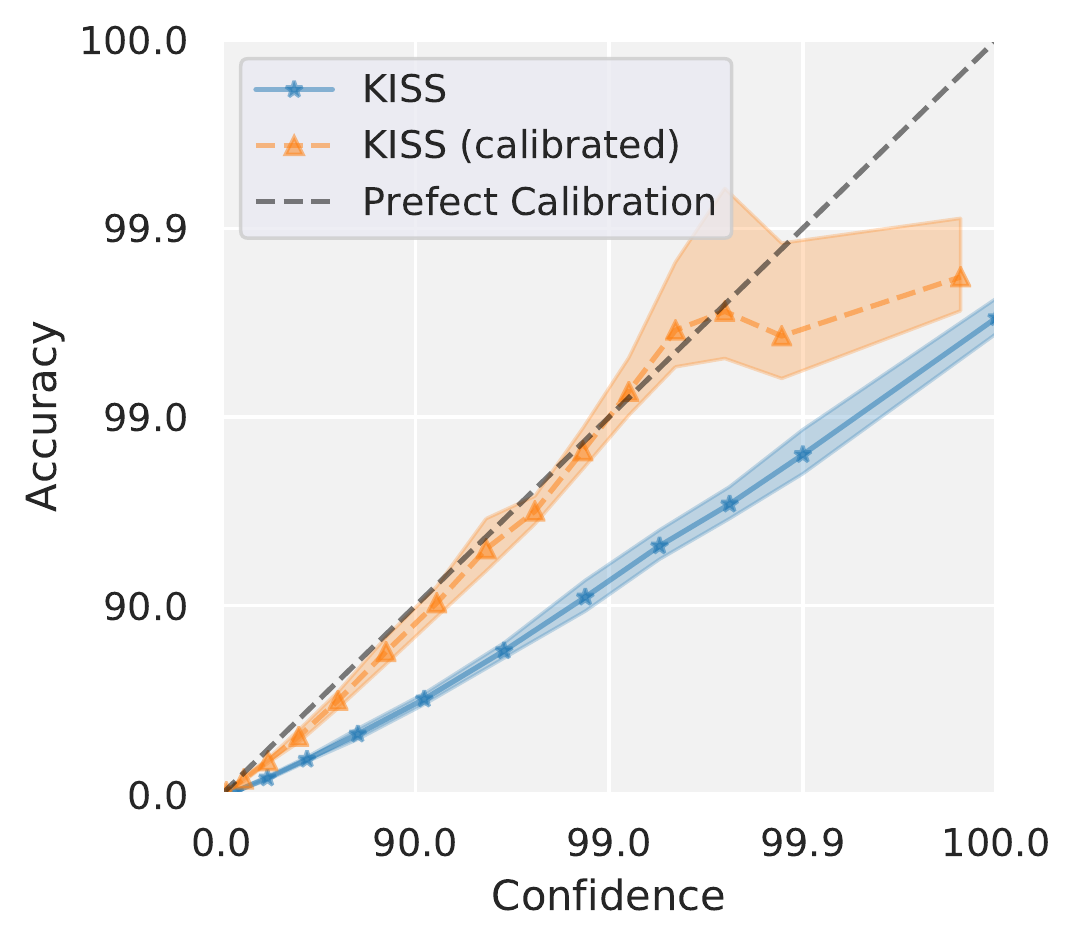}
    \hspace{-12pt}
    \vspace{-10pt}
\end{center}
   \caption{\textbf{Reliability Diagrams \cite{degroot1983comparison}}: 
   (i) AttDec --  a variant of ~\cite{Baek2019clova} with an attention decoder, 
   (ii) CTCDec -- a variant of ~\cite{Baek2019clova} with a CTC decoder, 
   (iii) SCATTER \cite{Litman_2020_CVPR},
   (iv) SEED \cite{qiao2020seed},
   (v) BiLSTM-AttDec --  a variant of ~\cite{Baek2019clova} with an attention decoder and a BiLSTM sequential feature encoder,
   (vi) BiLSTM-CTCDec --  a variant of ~\cite{Baek2019clova} with an attention decoder and a BiLSTM sequential feature encoder,
   (vii) ASTER \cite{Bai2018aster} and
   (viii) KISS \cite{bartz2019kiss}. We calibrate using T-scaling coupled with an equal bin size ECE objective applied to the word-level scalar confidence scores. The accuracy here is measured w.r.t exact word match.
   The figure shows accuracy vs. confidence plotted for equally-sized confidence bins, before and after calibration. Over-confidence can be observed for STR models, where the confidence of the model is higher than the expected accuracy.
  }
\vspace{-0.3cm}
\label{fig:reliability_diagrams_supp}
\end{figure*}

Figure \ref{fig:reliability_diagrams_supp} extends the reliability plots from the paper to encompass all studied models. As discussed in the paper, diagrams are presented on a log-log scale due to the natural distribution of confidences.

\section{NLL}
\label{sec:nll_supp}
The Negative Log Likelihood objective is commonly used during confidence calibration. NLL is defined as:

\begin{equation} 
\mathcal{L}=-\sum_{i=1}^{n} \log \left(\hat{\pi}\left(y_{i} \mid \mathbf{x}_{i}\right)\right),
\label{eq:nll}
\end{equation}

where the estimated probability $\hat{\pi}$ for the ground truth label $y_i$ given the sample $x_i$ is formulated as

\begin{equation*}
\hat{\pi}\left(y_{i} \mid \rvx_i\right) = \softmax \left( \rvz_{i}\right)^{(y_i)}.
\label{eq:nll-prob}
\end{equation*} 

Namely, for sample $i$, it is the value of the softmax output vector for logits vector $\rvz_i$, at the position corresponding to the correct prediction, $y_i$. The NLL objective is commonly used in calibration of classification models \cite{on_calib_icml17}; however, it is not well suited for the case of sequential decoding. To demonstrate this we must first redefine classic NLL for the case of sequential predictors. We first define the sequence estimated probability as: 

\begin{equation}
    \hat{\pi}\left(y_{i} \mid \rvx_i\right) = \prod_{j \in |\rvz_i|} \softmax(\rvz_{i,j})^{(y_i)}
    \label{eq:seq_estimation_prob}
\end{equation}

Here $\rvz_{i,j}$ signifies the output logit for the $j^{th}$ of the $i^{th}$ word, and $|\rvz_i|$ is the length of sample $i$
Taking Equation~\ref{eq:seq_estimation_prob} and plugging it into Equation~\ref{eq:nll} we derive the NLL objective over the space of sequence prediction probabilities.

Ideally, minimizing this objective would achieve good calibration at the sequence level, unfortunately this is not the case. By logarithm rules we obtain:

\begin{align*} 
\mathcal{L}&=-\sum_{i=1}^{n} \log \left( \prod_{j \in |\rvz_i|} \softmax(\rvz_{i,j})^{(y_i)}\right)=\\
&=-\sum_{i=1}^{n}\sum_{j \in |\rvz_i|} \log \left( \softmax(\rvz_{i,j})^{(y_i)}\right).
\end{align*}   

During calibration, the sequential objective function achieves the same minimum as Equation~\ref{eq:nll} and therefore essentially minimizes the per-character calibration error. 
\section{Calibration by Word Length All Results}
\label{sec:calib_word_length_supp}
\begin{figure*}[h!]
\begin{center}
ASTER
\includegraphics[width=0.27\textwidth]{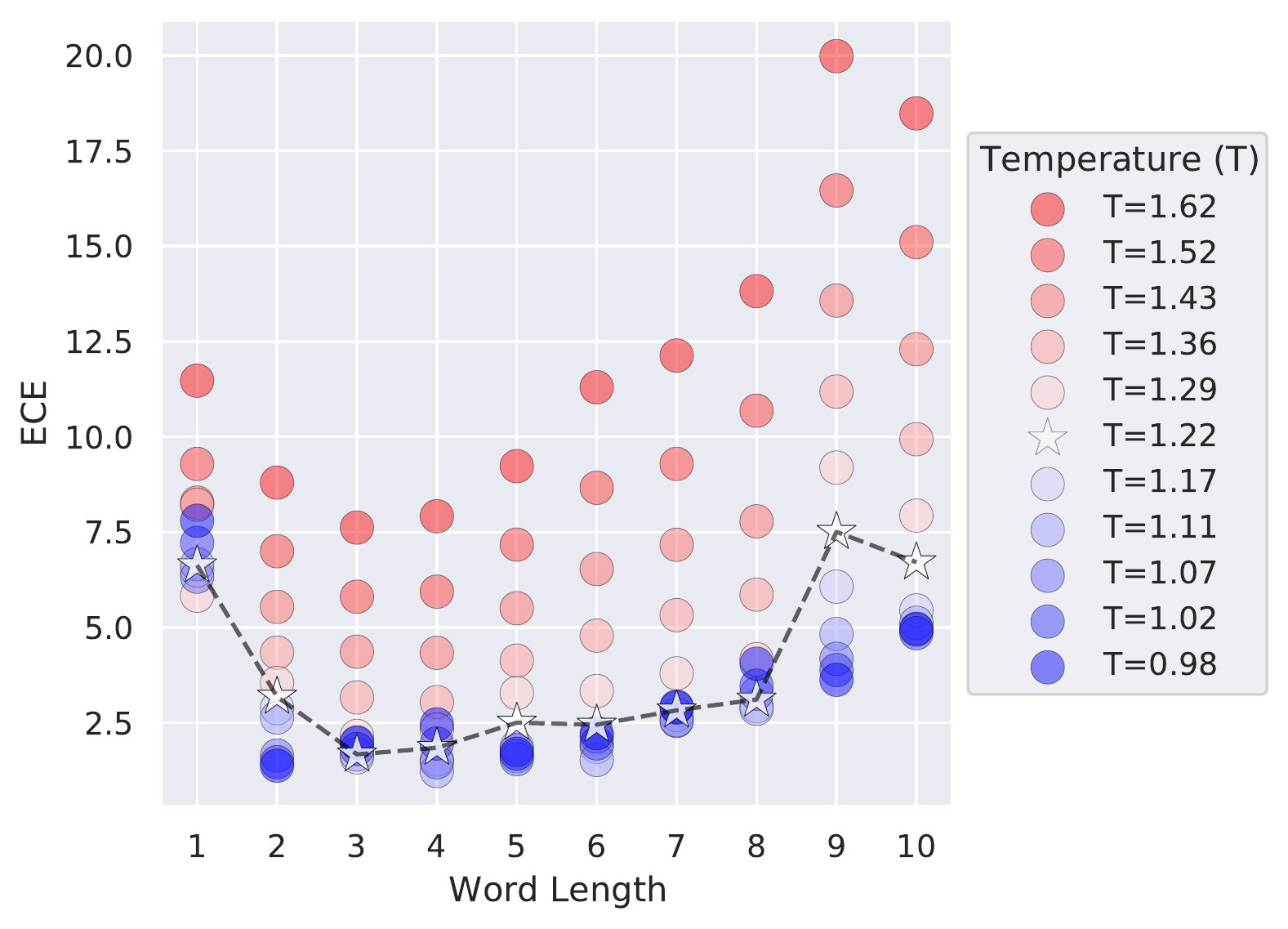}
SCATTER
\includegraphics[width=0.27\textwidth]{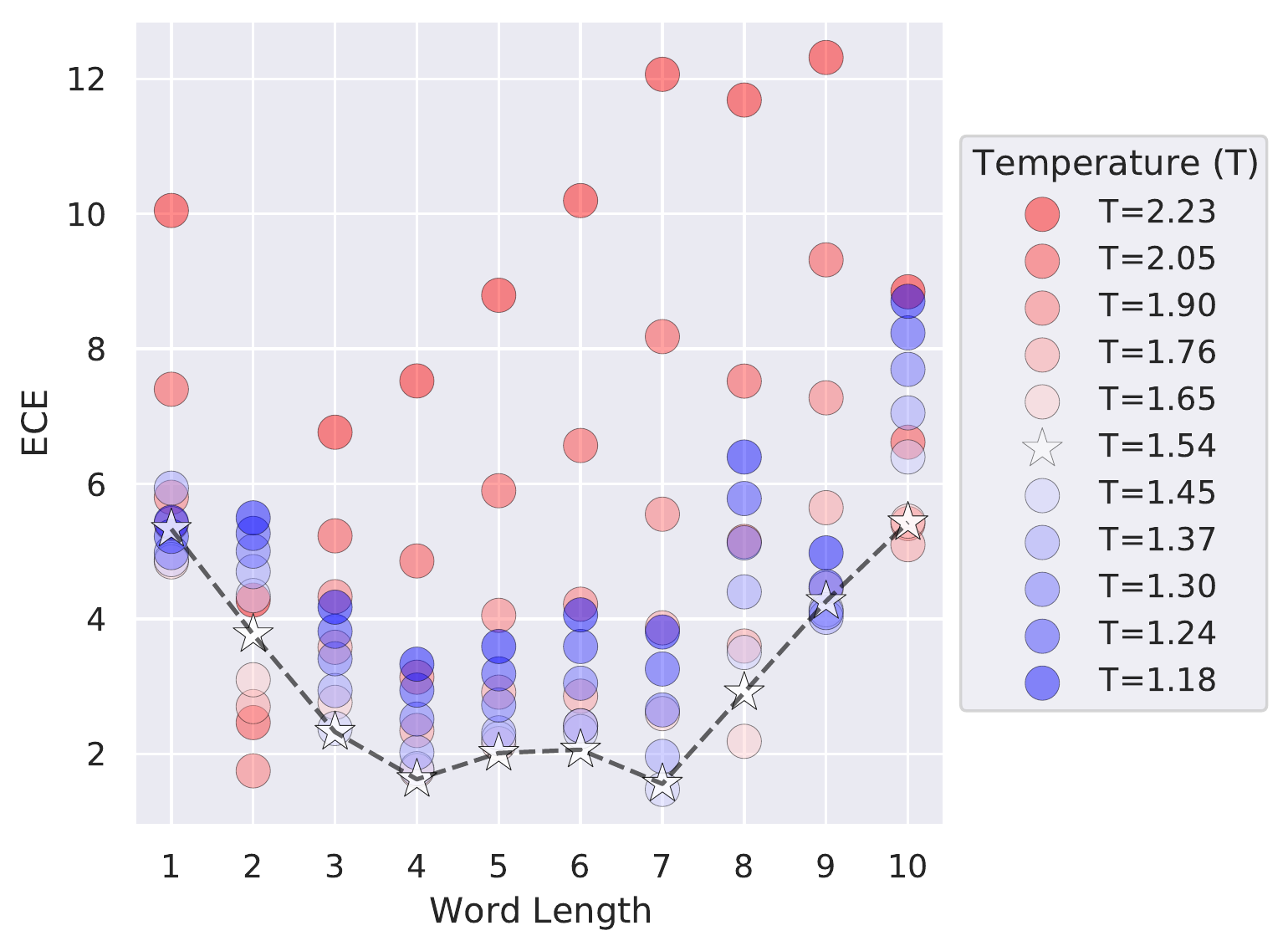}
SEED
\includegraphics[width=0.27\textwidth]{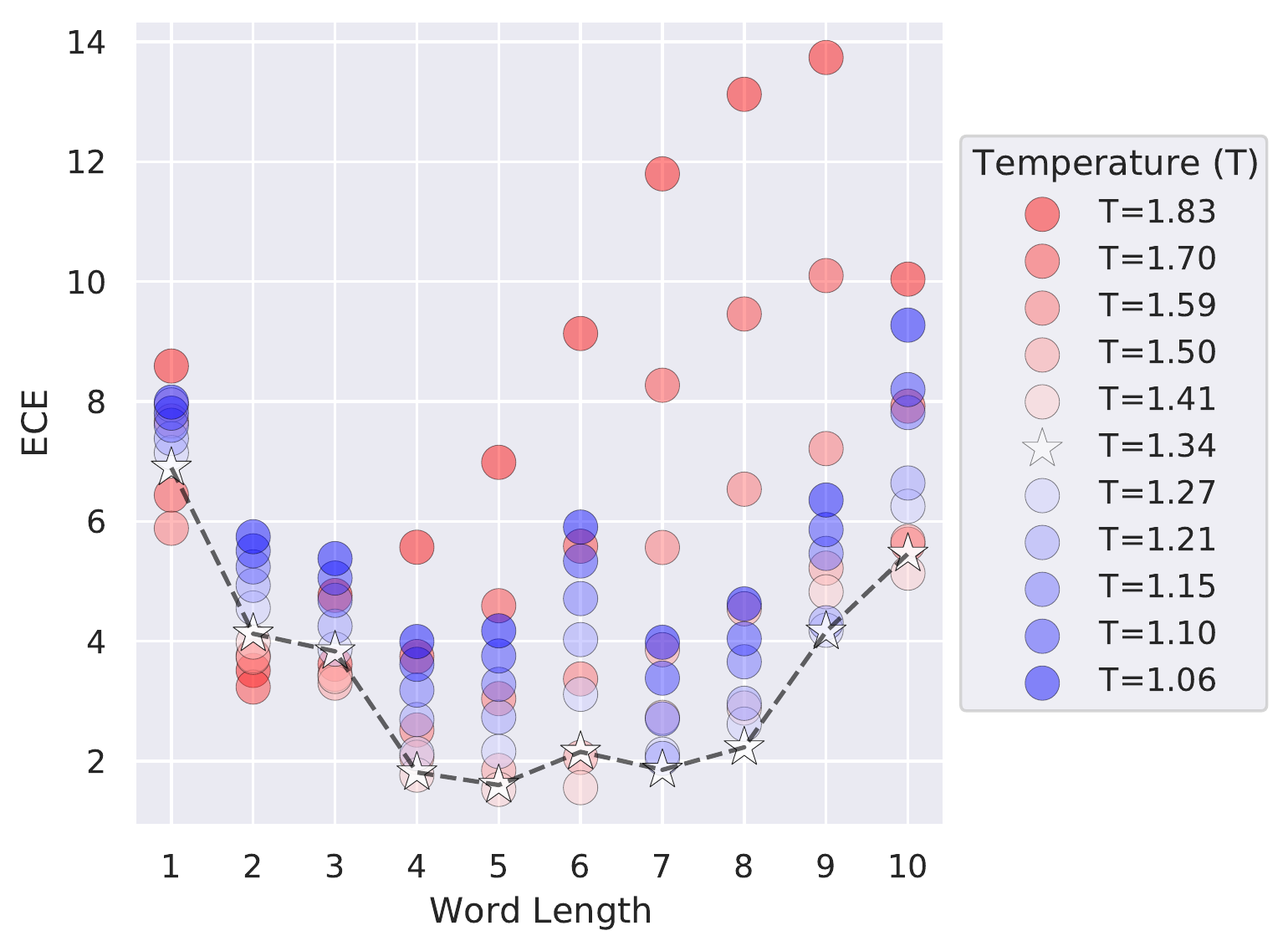}
CTC
\includegraphics[width=0.27\textwidth]{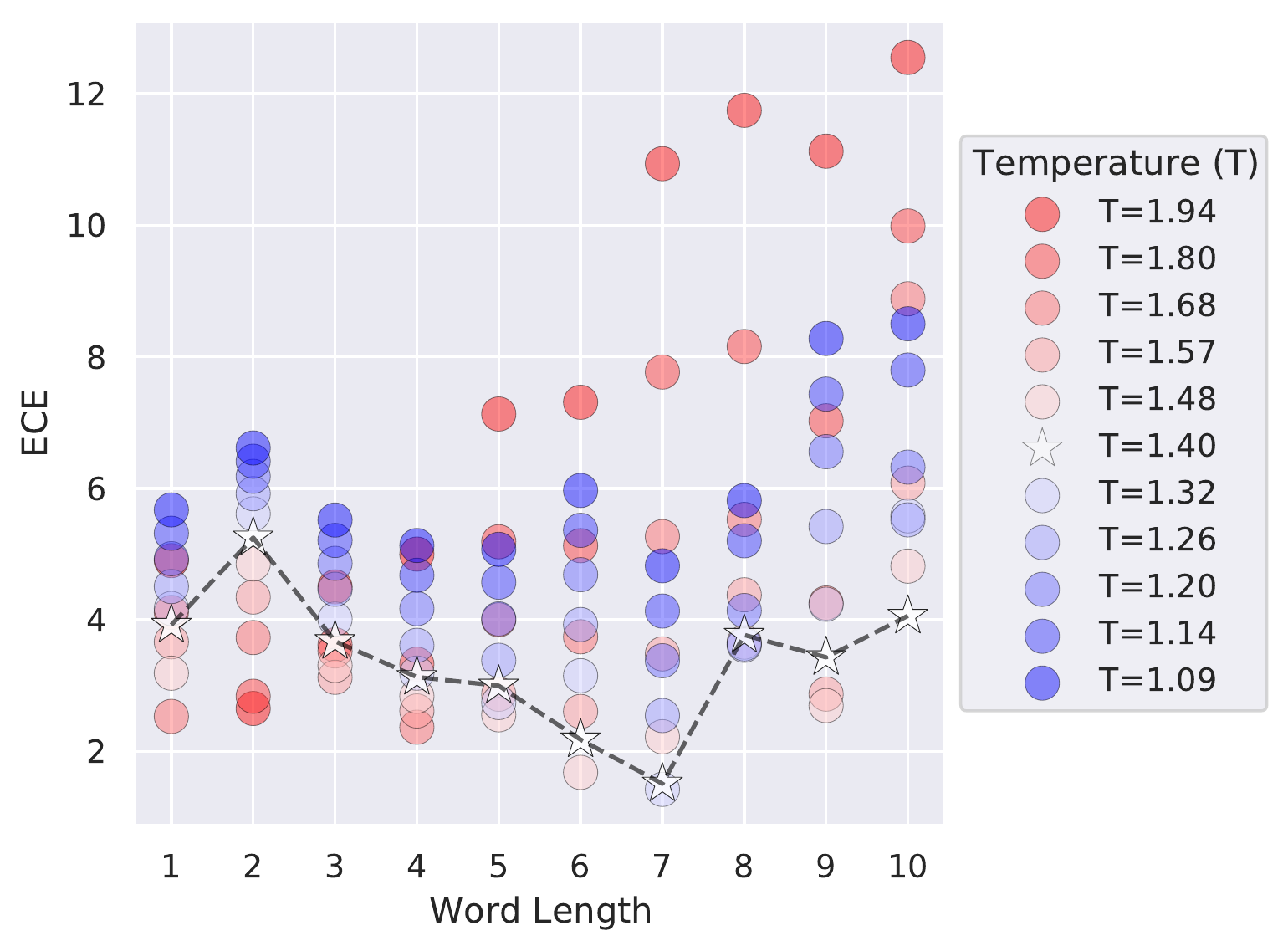}
BiLSTM CTC
\includegraphics[width=0.27\textwidth]{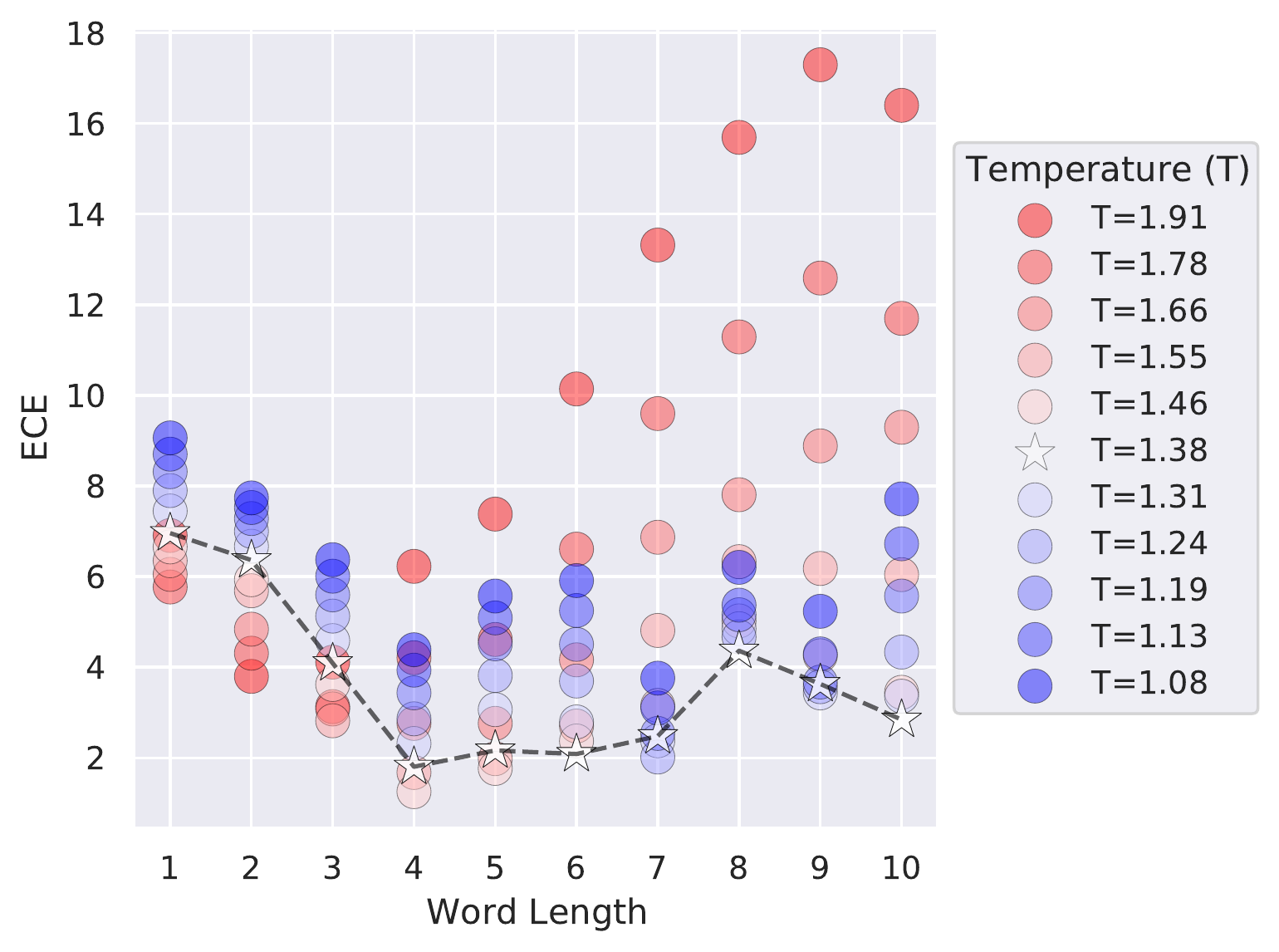}
\end{center}
   \caption{\textbf{ECE for different word lengths and temperatures}. Here, the optimal T-scaling temperature marked with a white star, warmer temperatures are marked with red circles and colder ones in blue. this figure extends the one presented in the paper to five different models depicting a similar pattern of robustness of the optimal temperature.}
   \label{fig:calib_by_length_sup}
\end{figure*}

We extend here the results presented in the main paper for the calibration by word-length experiment. Figure \ref{fig:calib_by_length_sup} presents ECE values measured for several temperatures surrounding the optimal calibrated temperature. The figure depicts this for five different studied methods, demonstrating the robustness of T-scaling to word length.
\section{Beam-Search Results}
\label{sec:beam-search_supp}
In Table \ref{tab:beam-search_supp} we present beam-search results for a beam width of $4$ broken down by dataset. Each model presents the calibrated ($\checkmark$) and non-calibrated ($X$) accuracy results on a per-dataset basis. As in the paper, results appear as a difference in accuracy with relation to a beam width of $1$ \ie no beam-search performed.
\begin{table*}[h!]
	\centering
\setlength{\tabcolsep}{3pt}
\begin{tabular}{l c c c c c c c c c}
\toprule
\textbf{Method}                         & \textbf{Calibrated} & \begin{tabular}[c]{@{}c@{}}\textbf{CUTE}\\ 288\end{tabular} & \begin{tabular}[c]{@{}c@{}}\textbf{IC03}\\ 867\end{tabular} & \begin{tabular}[c]{@{}c@{}}\textbf{IC13}\\ 1015\end{tabular} & \begin{tabular}[c]{@{}c@{}}\textbf{IC15}\\ 2077\end{tabular} & \begin{tabular}[c]{@{}c@{}}\textbf{IIIT5k}\\ 3000\end{tabular} & \begin{tabular}[c]{@{}c@{}}\textbf{SVT}\\ 647\end{tabular} & \begin{tabular}[c]{@{}c@{}}\textbf{SVTP}\\ 645\end{tabular} & 
\begin{tabular}[c]{@{}c@{}}\textbf{Average}\\ \end{tabular} \\

\midrule
 \multirow{2}{*}{Atten.~\cite{Baek2019clova}}& X          & 0   & 0      & +0.1       & +0.77       & +0.07         & +0.31 & +0.16 & +0.26    \\
                                & \large{$\checkmark$}          & +1.39   & +0.12      & +0.2       & +0.92       & +0.2         & +0.46 & +0.77 & +0.47    \\
                                \midrule

\multirow{2}{*}{BiLSTM\_AttDec~\cite{Baek2019clova}}       & X          & +0.35   & +0.23      & +0.2       & +0.24       & +0.07         & +0.77 & 0 & +0.2    \\
                                & \large{$\checkmark$}          & +0.7   & +0.23      & +0.3       & +0.53       & +0.13         & +0.77 & +0.47 & +0.35    \\
                                \midrule
\multirow{2}{*}{ASTER~\cite{Bai2018aster}}          & X          & +1.04   & +0.58      & +0.5       & +0.8       & +0.40         & +0.16 & +1.86 & +0.64    \\
                                & \large{$\checkmark$}          & +0.35   & +0.69      & +0.6       & +0.87       & +0.6         & +0.47 & +2.79 & +0.82    \\
                                \midrule
\multirow{2}{*}{SCATTER~\cite{Litman_2020_CVPR} }         & X          & -0.35  & 0      & +0.1       & +0.39       & +0.07         & +0.31 & +0.62 & +0.19    \\
                                & \large{$\checkmark$}          & 0   & 0      & +0.1       & +0.43       & +0.13         & +0.15 & +1.40 & +0.28    \\
                                \midrule
\multirow{2}{*}{SEED~\cite{qiao2020seed}}           & X          & +1.74   & 0      & +0.2       & +0.34       & +0.2         & +0.62 & +0.46 & +0.32    \\
                                & \large{$\checkmark$}          & +1.74   & +0.12      & +0.59       & +0.48       & +0.23         & +0.77 & +0.31 & +0.42  \\ 
\bottomrule
\end{tabular}
\caption{\textbf{Beam-Search Accuracy Results} broken down by dataset. Results are presented for calibrated ($\checkmark$) and non-calibrated ($X$) versions of beam-search. it can be seen the for the most part, calibration improves performance for each model and each dataset. Results are presented as difference from non-beam-search inference.}
\label{tab:beam-search_supp}
\end{table*}



\end{document}